\documentclass[twoside]{article}

\usepackage[accepted]{aistats2025}
\usepackage[round]{natbib}

\usepackage{amsmath}
\usepackage{amssymb}
\usepackage{mathtools}
\usepackage{amsthm}

\usepackage{dsfont}
\usepackage{xcolor}

\usepackage{subcaption}

\theoremstyle{plain}
\newtheorem{theorem}{Theorem}[section]
\newtheorem{proposition}[theorem]{Proposition}
\newtheorem{lemma}[theorem]{Lemma}
\newtheorem{corollary}[theorem]{Corollary}
\theoremstyle{definition}
\newtheorem{definition}[theorem]{Definition}
\newtheorem{assumption}[theorem]{Assumption}
\theoremstyle{remark}

\usepackage{hyperref}
\hypersetup{
	colorlinks=true,       %
	linkcolor=blue,        %
	citecolor=blue,        %
	filecolor=magenta,     %
	urlcolor=blue         
}

\begin{document}

\runningauthor{G\'{o}is, Mofakhami, P. Santos, Gidel, Lacoste-Julien}

\twocolumn[

\aistatstitle{Performative Prediction on Games and Mechanism Design}

\vspace{-0.5cm}
\aistatsauthor{ Ant\'{o}nio~G\'{o}is$^{1}$ \And Mehrnaz Mofakhami$^{1}$ \And Fernando P. Santos$^{2}$}
\vspace{0.1cm}
\aistatsauthor{  Gauthier Gidel$^{*\, 1, 3}$ \And Simon Lacoste-Julien$^{*\, 1, 3}$ }
\vspace{0.1cm}
\aistatsaddress{$^{1}$Mila \& Universit\'{e} de Montr\'{e}al \And $^{2}$Informatics Institute, University of Amsterdam} 
\vspace{-0.7cm}
\aistatsaddress{$^{3}$Canada CIFAR AI Chair } 
\vspace{-0.1cm}

]

\begin{abstract}
  Agents often have individual goals which depend on a group's actions. If agents trust a forecast of collective action and adapt strategically, such prediction can influence outcomes
  non-trivially, resulting in a form of performative prediction. This effect is ubiquitous in scenarios ranging from pandemic predictions to election polls, but existing work has ignored interdependencies among predicted agents. As a first step in this direction, we study a collective risk dilemma where agents dynamically decide whether to trust predictions based on past accuracy. As predictions shape collective outcomes, social welfare arises naturally as a metric of concern. We explore the
  resulting interplay between accuracy and welfare, and demonstrate that searching for stable accurate predictions can minimize social welfare with high probability in our setting. By assuming knowledge of a Bayesian agent behavior model, we then show how to achieve better trade-offs and use them for mechanism design.

\end{abstract}

\section{INTRODUCTION}
Recent frameworks such as performative prediction study how predictions influence the distribution they aim to predict \citep{hardt2023performative_pastfuture}.  These have focused on accuracy for one predictor and independent predicted agents: a spam producer changes its content to fool a spam classifier \citep{adversarial2004dalvi,hardt2016strategic}, or one loan applicant adapts to improve its credit score ignoring adaptation by others \citep{perdomo2020performative}.

Performative prediction typically considers a larger set of independent data points, but interdependencies among predicted agents have been abstracted by the literature, and are not explicitly modeled. Existing extensions to multi-agent performative prediction focus on multiple \emph{predictors} \citep{narang_ratliff2022multiplayer}, but not on multiple interdependent \emph{predicted}. However a plethora of examples exists requiring a collective scale among predicted. In election polls, the aggregate prediction of voters' behaviour influences individuals' actions \citep{simon1954bandwagon,blais2006polls}. Other times prediction is not directly about aggregate behaviour, but about its consequence. In pandemic modelling, forecasts of disease spread can change real spread, as people infer the consequences of collective action on the predicted outcome. Additionally, potentially accurate pandemic forecasts which were not observed due to performative effects can erode public trust in future predictions \citep{van2021pandemic_perform}. Collective action can also be shaped by predictions of a climate disaster, and road traffic by estimated travel time. In financial markets, price is an aggregate consequence of actions steered by predictions \citep{soros1987alchemy}. Such self-fulfilling prophecies may have actually deeply harmed society in cases such as the British pound collapse in 1992 \citep{naef2022exchange}, illustrating how welfare can be affected in multi-agent settings. Overall, these examples motivate a framework where the predicted population is interdependent, has a utility that only depends indirectly on predictions, trust varies with past accuracy, and welfare becomes an additional metric of concern.
To model a social dilemma such as pandemic containment or cooperation for climate change, here we propose the first setting with inherent interdependence among predicted agents\footnote{Code available at \url{https://github.com/antoniogois/performative-games}}. Agents play a cooperation game whose outcome depends locally on others' actions, and decisions are influenced by predictions of individuals' actions. Moreover, we allow for spatial structure among agents through a graph, representing social connections or geographic locations of common goods. Predictions are provided about individuals' actions, from which each one infers its group's expected behaviour.
Each agent updates a Bayesian trust variable based on past accuracy, which determines to what extent its actions are influenced by predictions.

Social welfare naturally emerges as a metric of interest in addition to accuracy. Welfare %
depends on the population's current actions, which are a consequence of predictions. %
Agents' trust links accuracy and welfare, by limiting the influence of inaccurate predictions in the long-term. As in most game theoretical settings, there is a tension between individual and collective interests, rendering high-welfare states possibly unstable, while accuracy maximization remains oblivious to this issue. Despite this, the broad goal of existing frameworks has been to minimize a single risk under performativity, which typically represents 
accuracy \citep{perdomo2020performative}. In our examples, welfare is only influenced indirectly by the predictor, making it unsuitable for standard performative prediction or supervised learning. Optimizing this kind of risk raises additional challenges \citep{miller2021outside_echo}, rendering the problem hard to solve in general, without tailoring solutions to problem instances with specific structures.

Despite a gap in the literature, having a secondary risk that does not depend directly on predictions is common when predicted agents are not independent. Predictions of pandemic growth or climate change can inform public policy, and become performative if risk is successfully reduced. In financial markets, predictions may aim at maximizing profit instead of accuracy. In elections, each candidate wishes to push the forecast whose collective reaction will benefit them the most. Even if a neutral entity wishes to deploy an accurate election poll, its performative effect may have strong unintended consequences \citep{ westwood2020confidence_elections}.

This highlights an aspect of mechanism design, which exists even when ignored. While deliberately deploying a wrong prediction is not an ethical option, there may be multiple possible realities that can be induced \citep{hardt2022performativepower} — therefore different predictions may be equally correct. Providing a snapshot of pre-prediction reality 
may be a way out of this dilemma, but can be wrongly interpreted as a prediction of post-prediction reality. The choice of how many snapshots to provide before action will also influence arbitrarily the outcome. Providing counterfactual scenarios for different population responses can improve transparency, but may risk an overly complex message being ignored by the predicted \citep{van2021pandemic_perform}. Our work illustrates this problem and difficult choices that arise from it, through the following contributions:
\begin{itemize}
    \item We propose a novel performative setting where the predicted population is inherently interdependent. For this, we adapt the Collective Risk Dilemma \citep[CRD;][]{milinski2008collective_risk} to include a centralized predictor, together with a trust variable that reduces performative effects over the predicted, as accuracy decreases.
    \item %
    We show how a second risk, welfare, appears when agents have goals that do not depend directly on the prediction. 
    We also propose a \emph{trust} model that ties welfare to accuracy, as 
    it hinders reaching states that require inaccurate predictions.
    \item We show in our setting how repeated risk minimization (RRM), a realistic algorithm to maximize accuracy under performativity, can accidentally minimize welfare with high probability over the initial predictions.
    \item We propose a method to learn how to use predictions as a mechanism to increase cooperation and reach higher welfare states. We show experimentally that this mechanism can achieve higher welfare in exchange for a drop in accuracy. 
\end{itemize}

\section{A MODEL FOR PREDICTING COLLECTIVE ACTION}

We are interested in game-theoretic scenarios where a population is interdependent and possibly influenced by predictions of collective behaviour. %
To that end, we propose a model where subgroups from a larger population interact simultaneously in a given round, drawing inspiration from evolutionary game theory on networks \citep{smith1982evolution, ohtsuki2006simple}. Given a graph $\mathcal G=(V,E)$, for any agent $i$, its group is composed of $i$ (itself) and its neighbors in the graph $\mathcal{N}(i)$. For one round of the game, agents simultaneously select an action, and each agent $i$ receives a payoff $\pi_i(y_i,y_{\mathcal N(i) })$ depending on its own action $y_i$ and on neighbours' $y_{\mathcal N(i)}$. The game repeats for $R$ rounds.

To choose $\pi_i$, we focus on CRD, suitable to study mechanism design \citep{gois2019reward}. Each round requires a critical mass of cooperators to achieve success and prevent collective losses. This may represent the protection of common natural resources, the immunity of a partially vaccinated group, and the collective development of tools like Wikipedia or Linux, among many others. If the fraction of cooperators remains below a threshold $T$, collective success is not achieved and everyone risks losing their endowment with probability $r$. Each agent chooses whether to defect ($y_i=0$) or cooperate at a cost ($y_i=1$), with payoffs below:

\begin{definition}
    (Defector's payoff) Let $\mathds{1}[\cdot]$ be the indicator function. $k_i=\sum_{j\in\mathcal{N}(i)\cup \{i\}}y_j$ is the number of cooperators in agent $i$'s group. Given an initial endowment $B$, $k_i$ cooperators in a group of size $M_i$, threshold $T$ where $0\leq T\leq 1$, and risk $r$, where $0\leq r \leq 1$, the expected payoff of defector $i$ is
    \begin{align}\label{eq:defector}
        \pi_{{D_i}}(k_i) = B\cdot(&\mathds{1}[k_i\geq\lceil TM_i\rceil] \notag \\
        &+(1-r)\mathds{1}[k_i<\lceil TM_i\rceil])
    \end{align}
\end{definition}

In words, below threshold there is a disaster with probability $r$, while over the threshold each agent gets $B$.

\begin{definition}(Cooperator's payoff) Given a cost $cB$ of cooperating, where $0\leq c\leq 1$, the payoff of cooperator $i$ is

\vspace{-\baselineskip}
\begin{align}
    \pi_{{C_i}}(k_i) = \pi_{{D_i}}(k_i) - cB
\end{align}
\end{definition}

A CRD is used as payoff function $\pi$ for all agents, using the same threshold value $T$ and unique $M_i$'s given by $\mathcal G$. This leads to partially aligned incentives — each agent $i$ gains from preventing a disaster where $\frac{k_i}{M_i}<T$, but would rather avoid incurring cost $cB$ of cooperating to increase $k_i$.

For one round of CRD with $c<r$ and one single group (where $\mathcal{G}$ is a fully-connected graph $\mathcal{G}_f$), the Nash equilibria \footnote{A state is a Nash equilibrium if no agent can unilaterally change her action to improve her payoff $\pi$.} are for everyone to defect (sub-optimal) or to have exactly $\lceil TM_i\rceil$ cooperators (Pareto optimal \footnote{A state is Pareto optimal if there is no other state that improves one's $\pi$ without lowering another's $\pi$.}). The challenge is in coordinating a group towards the Pareto optimal Nash, which doesn't happen spontaneously in the real world \citep{milinski2008collective_risk}. 

\subsection{Agent Model}

We model agents as computing a best-response, given expectations of other individuals' actions. To nudge behaviour, a predictor provides predictions of the population actions. Alternatively to correlated equilibria \citep{aumann1974correlated_eq} we provide a public signal, which agents can choose to trust or not. Since this signal is learned from global observations of the whole population (and not just locally) it has the potential to bring additional information to agents. We assume agents observe a public prediction of others' actions, but stop trusting it if it is inaccurate. More specifically, they follow a Bayesian update to compute the probability of trusting the prediction. Agent $i$ has two competing explanations for each neighbour $j$'s behaviour — the external prediction $\hat y_j$ and an internal expectation $\alpha_{i,j}$. Both $\hat y_j$ and $\alpha_{i,j}$ are Bernoulli parameters that estimate a hypothetical true  $\mathbb P(y_j=1)$.
$\mathcal L_i(\boldsymbol{\hat y}_t, \boldsymbol{y}_t)$ is the likelihood of observing $\boldsymbol{y}_t$ given parameter $\boldsymbol{\hat y}_t$, and $\tau_{t-1,i}$ acts as the prior for $i$.
The posterior probability $\tau_{t,i}$ of $i$ trusting the external predictor in timestep $t$ is:

\vbox{
\begin{align}
    \tau_{t,i}&= \mathbb P(\text{\scriptsize trust}|
    \boldsymbol{\hat y}, \boldsymbol{y}
    ) \notag \\
    &= \frac{\tau_{t-1,i}\mathcal L_i(\boldsymbol{\hat y}_t, \boldsymbol{y}_t)}
{\tau_{t-1,i}\mathcal L_i(\boldsymbol{\hat y}_t, \boldsymbol{y}_t) + (1-\tau_{t-1,i})\mathcal L_i(\boldsymbol\alpha_i, \boldsymbol{y}_t) }
\end{align}
}

with $\mathcal L_i(\boldsymbol {\hat y_t}, \boldsymbol{y}_t):=\underset{j\in \mathcal{N}(i)}{\prod}\hat y_{j,t}^{y_{j,t}}(1-\hat y_{j,t})^{1-y_{j,t}}$.

Given the expectation of others' actions, $i$ can compute a rational utility-maximizing action. As long as $c<r$, it is rational for $i$ to cooperate if and only if $\sum_{j\in \mathcal{N}(i)}y_j=\lceil TM_i\rceil-1$. In words, $i$ cooperates when it is the only missing cooperator required to overcome the threshold in its group. Given probability $\boldsymbol\theta_{\mathcal{N}(i)}=\theta_{1...M_i-1}$ of each neighbour of $i$ to cooperate, a Poisson binomial distribution $g(T_i|\boldsymbol\theta_{\mathcal{N}(i)})$ gives us the aggregate probability of having $T_i=\lceil TM_i\rceil-1$ cooperators in the group. \texttt{BestResponse} is then $\arg\max_{y_i} \mathbb E_{y_{\mathcal{N}(i)}\sim g(\boldsymbol\theta_{\mathcal{N}(i)})}[\pi(y_i, y_{\mathcal{N}(i)})]$.%

\begin{proposition}
    (Best-response under competing models) Given two competing models ($\boldsymbol{\hat y}$ and $\alpha$) that explain the population's behaviour, then $\textnormal{\texttt{BestResponse}}_i(\boldsymbol{\hat y}_{t+1,\mathcal N(i),\theta};\tau_{t,i})$ is to cooperate if $\tau_i g(T_i|\hat y_{j\in \mathcal{N}(i)})+(1-\tau_i)g(T_i|\alpha_{i,j\in \mathcal{N}(i)})>\frac{c}{r}$, and defect otherwise (Appendix~\ref{app:best_response})
\end{proposition}

\subsection{Performative Prediction in Collective Action}
\label{sec:perf_pred}

\citet{perdomo2020performative} propose the framework of performative prediction, and analyze RRM, an algorithm which retrains a model after each distribution shift. More formally, they assume a mapping $\mathcal D(\theta)$ from a predictor parameter $\theta$ into a distribution $\mathcal D$. Considering this effect of $\theta$, performative risk becomes:

\begin{align}
    \text{PR}(\theta)=\underset{Z\sim\mathcal D(\theta)}{\mathbb E} \ell(Z;\theta)
\end{align}

As a baseline to optimize $\text{PF}(\theta)$ without knowledge of $\mathcal D(\theta)$, they suggest RRM as a natural heuristic — to repeatedly minimize risk using the distribution obtained from the previous model deployment:

\begin{align}
    \theta_{t+1} = \underset{\theta}{\arg\min} \underset{Z\sim\mathcal D(\theta_t)}{\mathbb E} \ell(Z;\theta)
\end{align}

We now map their setting to our framework. We have a prediction $\boldsymbol{\hat y}_t\in [0,1]^{|V|}$ of each agent's probability of cooperating in time-step $t$, starting at $t=1$. A trust variable $\tau_{t,i}\in [0,1]$, starting at $t=0$, determines how much agent $i$ trusts predictions, and therefore by what extend its next action $y_{t+1,i}$ is influenced by $\boldsymbol{\hat y}_{t+1}$. 
Finally, we have:

\begin{multline}
    \mathcal D(\theta; \tau)=\{\mathds{1}[y_{t+1,i}=\\
    \texttt{BestResponse}_i(\boldsymbol{\hat y}_{t+1,\mathcal N(i),\theta};\tau_{t,i})]:i\!=\!1,...,|V|\}\\
\end{multline}

yielding a deterministic distribution.

The missing ingredient to apply RRM is which risk to use. We first consider accuracy maximization, in the simplified case of a static \mbox{$\tau=1$}, a single-round game, and a new deployment of 
$\boldsymbol{\hat y}$
after each round. %
The learner initializes $\hat y_1\in[0,1]^{|V|}$, after which each agent $i$ computes deterministically $y_{1,i}=\texttt{BestResponse}_i(\boldsymbol{\hat y}_{1,\mathcal N(i),\theta};\tau_{0,i})$, corresponding to $y_{1,i}\sim\mathcal D(\hat y_{1,\mathcal N(i)}; \tau_{0,i})$. Unaware of its own influence on $\boldsymbol{y}$, an accuracy-maximizing learner minimizes risk by deploying $\hat y_{2,i}=y_{1,i}$, staying in the discrete domain $\hat y_t\in\{0,1\}^{|V|}$ from time-step $t=2$ onwards.
Note we will not restrict ourselves to RRM or to $\tau=1$, but it will be the starting point of our analysis.

Accuracy fits naturally the definition of performative risk: $\text{PR}(\theta):=\mathbb E_{Z\sim\mathcal D(\theta)}\ell(Z;\theta)$. On the other hand, welfare is a metric that only depends on the actions taken by the population: $\mathbb E_{Z\sim\mathcal D(\theta)}\ell'(Z)$. Minimizing such a loss raises optimization challenges, since $\theta$ only influences risk through $\mathcal D(\theta)$.

\section{MODEL DYNAMICS}
\label{sec:theoretical_dynamics}

\begin{figure}[t]
\vskip 0.2in
\begin{center}
\centerline{\includegraphics[width=\columnwidth]{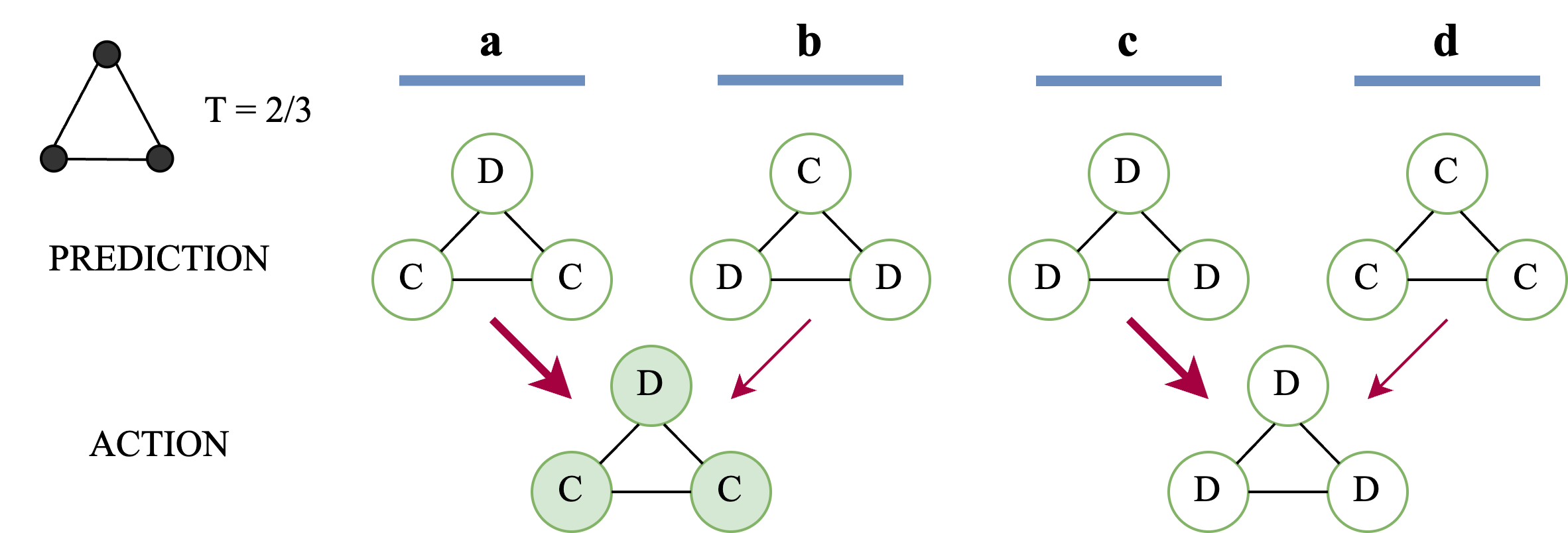}}
\caption{Dark nodes have achieved success, and thick arrows are self-fulfilling prophecies. Both a) and c) are self-fulfilling prophecies where accuracy is maximized, therefore an accuracy maximizer is indifferent between them. However, in a) full success is achieved, but in c) all fail. b) also maximizes group success but at the expense of 0\% accuracy.}
\label{fig:3-node clique}
\end{center}
\vskip -0.2in
\end{figure}

To gain understanding of our model's behaviour, here we study the impact of different components in isolation. We explore two opposite scenarios: blindly trusting a predictor ($\tau=1$), in particular the impact of $T, \mathcal G$ and RRM, and behaviour in the absence of a predictor ($\tau=0$). We then show oscillatory behaviour for a particular $\mathcal G, T$, when $\tau$ is dynamic.

\subsection{Threshold and Topological Constraints}
\label{sec:simple_envs}

We adopt the following assumption:

\begin{assumption}
(Simple controllable setting) a) agents have a fixed trust $\tau=1$ ignoring their internal beliefs $\alpha$, and b) predictions are binary: \mbox{$\boldsymbol{\hat y_t}\in \{0,1\}^{|V|}$.}
\label{ass:controllable}
\end{assumption}

Let a \textit{self-fulfilling prophecy} be when $\forall i, y_i=\hat y_i$. Assuming binary predictions is useful in this definition, since $y$'s need to match $\hat y$'s. Removing the interference of internal expectations $\alpha$ by having \mbox{$\tau_0=1$}, predictions become static: $\boldsymbol{\hat y_t}=\boldsymbol{\hat y}$. With full trust guaranteed, there is no need to balance between trust and other goals through time. Under Assumption~\ref{ass:controllable}, if agents are never indiferent between actions, predicting a strict Nash equilibrium is sufficient and necessary to have a self-fulfilling prophecy (i.e. $\forall i, \texttt{BestResponse}(\hat y_{\mathcal N(i)})=\hat y_i$).

Whether there is a Nash equilibrium that maximizes welfare determines whether the predictor must compromise accuracy to maximize it. %
Note that, as long as $\forall i,\lceil TM_i\rceil>1$, \textit{all-defecting} is always a self-fulfilling prophecy. 
Using Assumption~\ref{ass:controllable}, the topology of $\mathcal G$ and threshold $T$ become the only constraints determining whether a given system state is attainable.

\begin{figure}[t]
\vskip 0.2in
\begin{center}
\centerline{\includegraphics[width=\columnwidth]{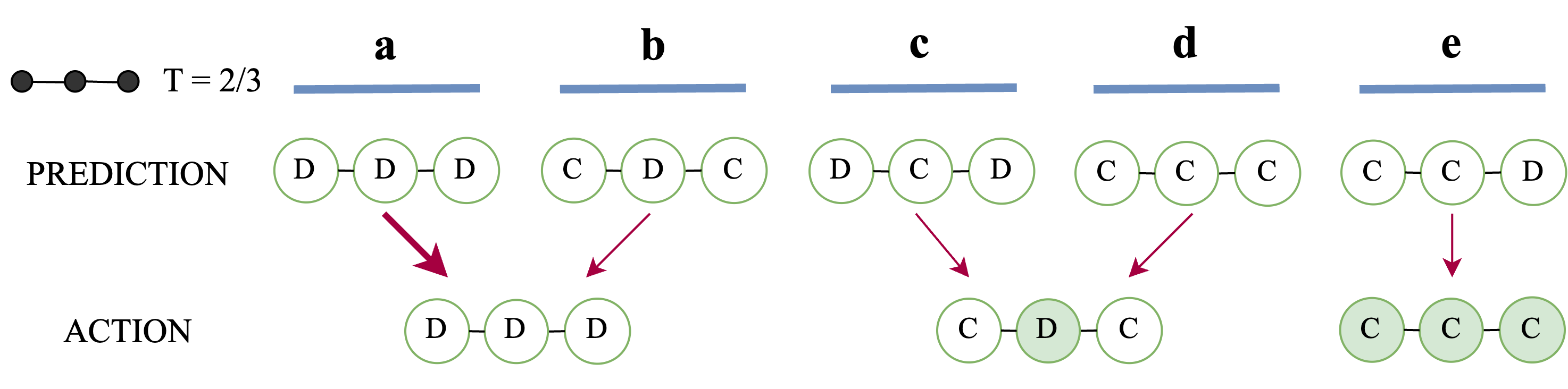}}
\caption{Dark nodes have achieved success, and thick arrows are self-fulfilling prophecies. 
Here there is no self-fulfilling prophecy which maximizes group success, forcing a trade-off between accuracy and group success. Only e) maximizes group success, but the center node regrets having cooperated. Note that, with $T=\frac{2}{3}$, groups of size $M_i=2$ require both agents to cooperate.}
\label{fig:3-node chain}
\end{center}
\vskip -0.2in
\end{figure}

\begin{proposition}
    (Sufficient conditions for self-fulfilling success) Let ``full success" be the setting where $\forall i, \frac{k_i}{M_i}\geq T$. Under Assumption~\ref{ass:controllable} and $c<r$, each of the following is a sufficient condition so that $\exists \, \boldsymbol{\hat y} \implies$ full success, where $\boldsymbol{\hat y}$ is a self-fulfilling prophecy:
    \begin{enumerate}
        \item $\mathcal{G}=\mathcal{G}_f$, where $\mathcal{G}_f$ is a clique or a fully-connected graph: Assume $\boldsymbol{\hat y}$ predicts a configuration with $k_i=\lceil TM_i\rceil$.  Since all agents share the same group, it is not possible for one agent to deviate from $\boldsymbol{\hat y}$ without lowering its $\pi$; 
        \item T=1: no agent can free-ride, since all are required to cooperate;
        \item  T=0: full success is guaranteed by default.
    \end{enumerate}
\end{proposition}

Figure~\ref{fig:3-node clique} illustrates the previous remarks, over a 3-node clique. a) and c) are Nash equilibria and self-fulfilling prophecies, while b) and d) are not. %
An accuracy maximizer would choose a) or c), while a welfare maximizer would choose a) or b). Here it is possible to maximize both quantities through a).

However, both goals may be at odds in other settings. In Figure~\ref{fig:3-node chain} there is no prediction that satisfies simultaneously an accuracy maximizer and a welfare maximizer. The only self-fulfilling prophecy is a), reaching full-defection. Only e) reaches full success, but since it is not a Nash it is not self-fulfilling. This is because the center node could have achieved success while defecting, but would have prevented success in groups of size 2. In general, full success is not always achievable, even if we do not require a self-fulfilling prophecy:

\begin{proposition}
\label{prop:nec_success}
    (Necessary condition for success) Under Assumption~\ref{ass:controllable}, a game must not obey simultaneously the three conditions below, otherwise full success is not attainable, even with a prediction that is not self-fulfilling (proof in Appendix~\ref{app:counter_example}):
    \begin{enumerate}
    \item graph $\mathcal{G}$ has a ``hub" node $H$ whose degree $M_H-1$ is higher than any of its neighbours: $\forall i\in\mathcal{N}(H): M_i<M_H$.
    \item Threshold $T\in[0,1]$ is set to $\frac{M_H-1}{M_H}$.
    \item $\forall i\in\mathcal{N}(H), \exists j\in\mathcal{N}(i):M_j<M_H$.
\end{enumerate}
\end{proposition}

\begin{figure}[!t]
    \centering
    \includegraphics[width=.95\columnwidth]
    {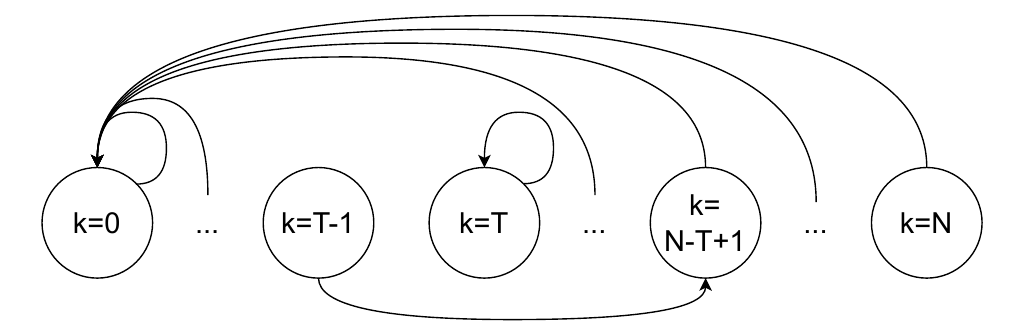}
    \caption{Markov chain describing the evolution of a population with $k$ out of $N$ cooperators, $\tau=1$ and $\mathcal{G}_f$ following RRM, for $N>2T-1,T\not=1$. (details in Appendix~\ref{app:low_welfare})}
    \label{fig:MPD_RRM}
\end{figure}

This shows how $T,\mathcal G$ can condition which states are reachable. The examples also illustrate how different predictions induce different realities in this model. As a consequence, seeking only high-accuracy predictions may inadvertently induce low-cooperation states, as the next section will show.

\subsection{Convergence to Low Welfare}
\label{sec:converge_low_welf}

We now show theoretically that repeatedly maximizing accuracy can minimize welfare with high probability, %
even when some regions of the optimization landscape allow to optimize for both metrics. As the previous section illustrates, the alignment between welfare and accuracy depends non-trivially on game parameters $T$ and $\mathcal G$, even in the simplified case of $\tau=1$. The choice of optimization algorithm introduces further complexity. Depending on it, even when it is possible to jointly optimize accuracy and welfare (e.g. Figure~\ref{fig:3-node clique}), an accuracy maximizer is not guaranteed to pick the best solution for both metrics.

Assume a predictor is trained to maximize accuracy through 
RRM, following \S~\ref{sec:perf_pred} with $\mathcal G=\mathcal{G}_f$. RRM becomes a Markov chain, where each state is defined only by the number of cooperators in the population, $k_t=\sum_i y_{t,i}$ in the fully-connected case. The initialization of $\boldsymbol{\hat y}_1$ will determine which $k_1$ is induced. From there on RRM will imitate each player's previous best-response, to which players will best-respond until convergence or reaching a cycle. This is equivalent to synchronous best-response dynamics \citep{chellig2022best_response}. Interestingly, this prevents agents from overcoming the threshold with high probability.

\begin{theorem}
\label{the:convergence}
    (Convergence to low welfare) Assume $\forall i,\lceil TM_i\rceil>1$, and $\mathcal G=\mathcal{G}_f$. Predictions \mbox{$\boldsymbol{\hat y}_1\in\{0,1\}^{|V|}$} can be described by $\hat k_1=\sum_i \hat y_{1,i}$ since $\mathcal G=\mathcal{G}_f$. Let $\hat k_1\sim \text{Bin}(\theta,N)$ and \mbox{$\theta\sim\text{Beta}(\alpha,\beta)$}, where $N=|V|$. RRM always converges to a stable point in terms of $k_t$ in at most 2 steps, and $\mathbb P(k_t=0)\geq 1 - \mathcal O(1/N)   \,,\; \forall t\geq 2$. (proof in Appendix~\ref{app:low_welfare})
\end{theorem}

\begin{figure}[!t]
    \centering
    \includegraphics[width=.8\linewidth]{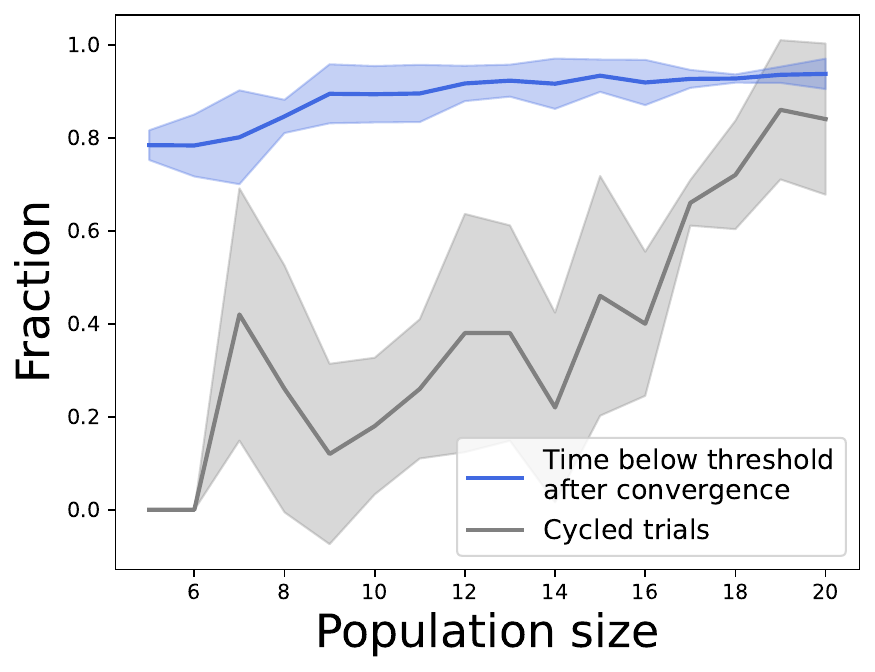}
    \caption{Average proportion of time spent below threshold, after convergence, and proportion of cycles. Each point is the average of 5 scale-free $\mathcal G$'s, with 10 random $\boldsymbol{\hat y}_1$ per $\mathcal G$, and $T=0.5$. Shaded areas are standard deviation among $\mathcal G$'s. Find histograms for time in each state in Appendix~\ref{app:low_welfare}.}
    \label{fig:n_cycled_pop_size}
\end{figure}

Empirically, convergence to low welfare holds for other kinds of $\mathcal{G}$'s. In particular, we study the configuration after convergence of scale-free networks, showing that most time is spent on low cooperation states, as size increases. Cycles become more frequent than a single steady state, in larger scale-free graphs (Fig~\ref{fig:n_cycled_pop_size}).

\subsection{Self-Organization without Predictor}
So far we have explored scenarios with full trust ($\tau=1$) where an external predictor dictates population actions, studying limitations on how much welfare can be obtained and behaviour under RRM. Now we focus on the opposite scenario, in the absence of a predictor.

For $\mathcal{G}=\mathcal{G}_f$ it is possible to compute the price of anarchy \citep{koutsoupias2009worst}, which is the ratio between the welfare of the best state and the welfare of the worst equilibrium. Defining welfare as $\text{Welf}(\boldsymbol{y})=\sum_i \pi_{y_i}(k_i)$, we obtain the price of anarchy $\text{PoA}=\frac{1-c\frac{\lceil NT\rceil}{N}}{1-r}\approx \frac{1-cT}{1-r}$. We are interested in regimes where $r>c$, and we note that PoA increases with higher $r$ and lower $c$, worsening the gap between best state and worst equilibrium.

\begin{figure}
    \centering
    \includegraphics[width=0.50\linewidth]{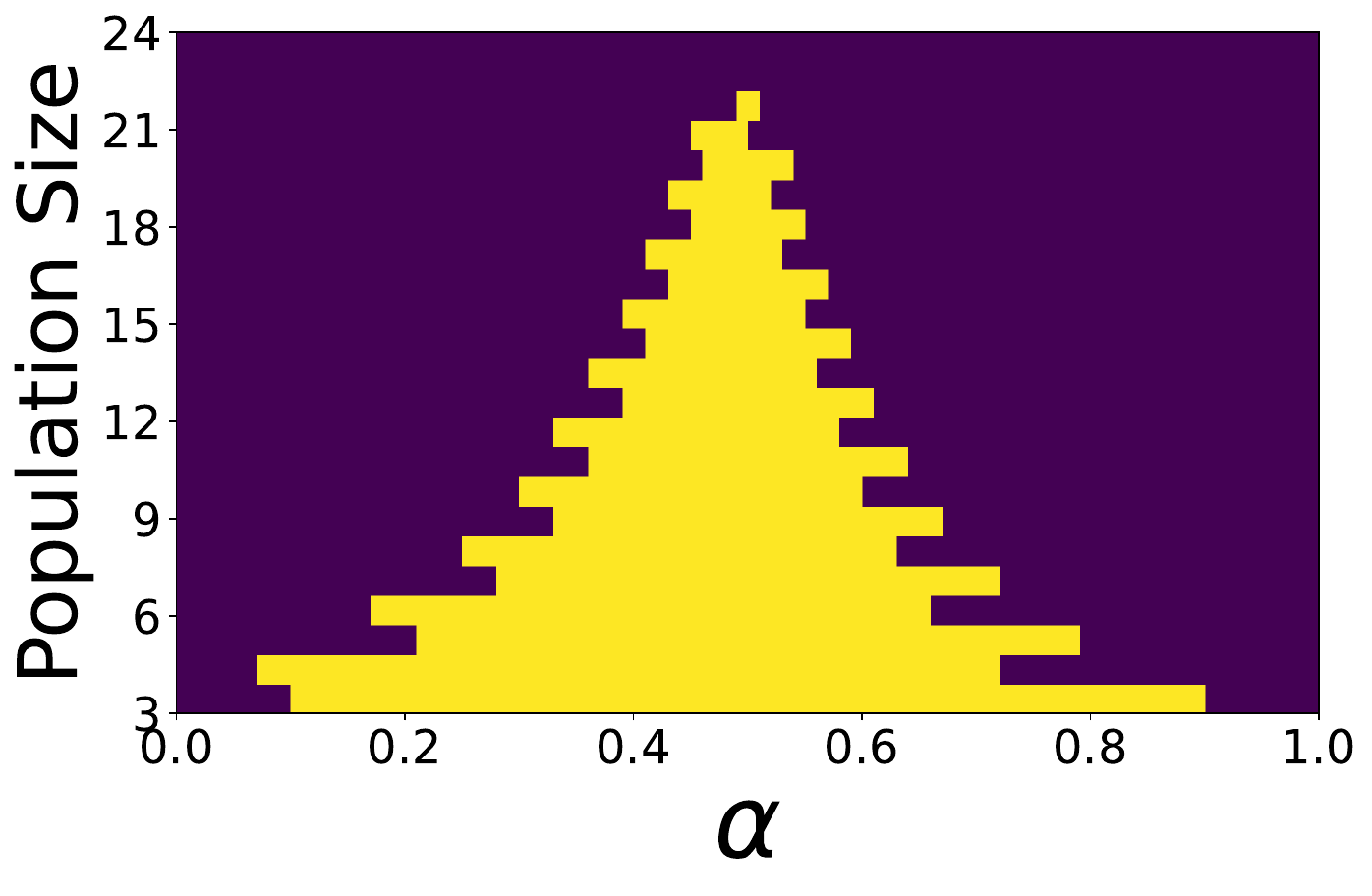}
    \hfill
    \includegraphics[width=0.45\linewidth]{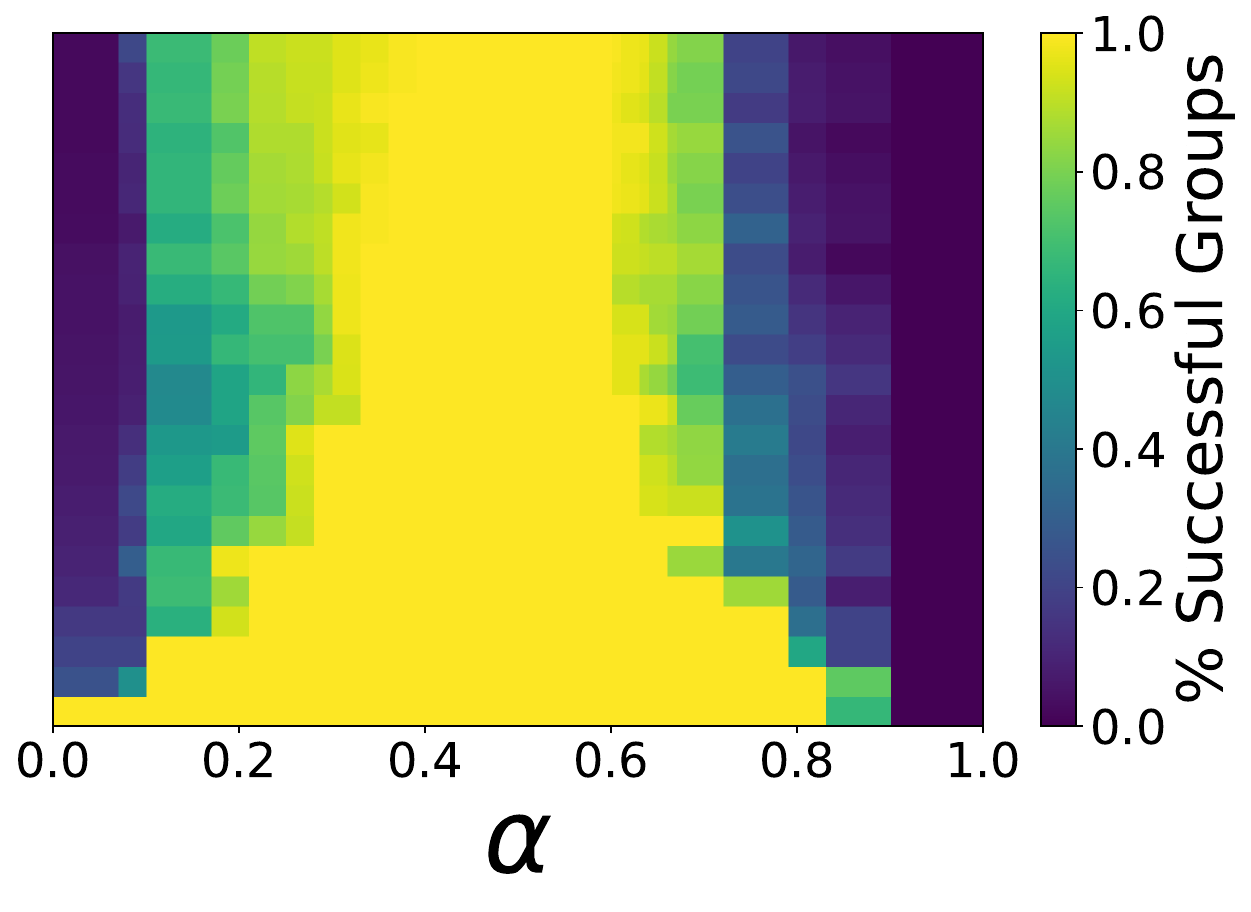}
    \caption{Proportion of successful groups with \mbox{$\tau=0$}, for a fully-connected graph (left) and scale-free networks (right) with varying $\alpha$ and population size N. $\frac{c}{r}=\frac{1}{6}$, $T=0.5$ and scale-free's average degree $m=2$.}
    \label{fig:alpha}
\end{figure}

PoA however does not consider to which equilibrium we arrive at without an external entity, focusing on the worst case. Figure~\ref{fig:alpha} shows results without trust ($\tau=0$), where agents behave according to their internal expectation $\alpha$ of others. For a fully-connected population of size $N$, each agent's best-response is to cooperate if $\text{Bin}(\lceil TN \rceil-1;\alpha,N)>\frac{c}{r}$. From this equation we see that as $\frac{c}{r}$ drops, PoA increases but agents choose to cooperate for a wider range of $T,N,\alpha$. For the remaining of the paper, we focus on regimes where high $\text{Welf}(\boldsymbol{y})$ does not happen spontaneously.

\subsection{Trust Dynamics}
\label{sec:trust_dyn}

When trust varies with accuracy, our system becomes stateful. The same prediction may induce different responses depending on how much each agent currently trusts predictions. Here we present a simplified analysis for the $\mathcal G,T$ described in Figure~\ref{fig:3-node chain}, assuming an infinite horizon discounted reward criterion.

\begin{figure}
    \centering
    \includegraphics[width=\linewidth]{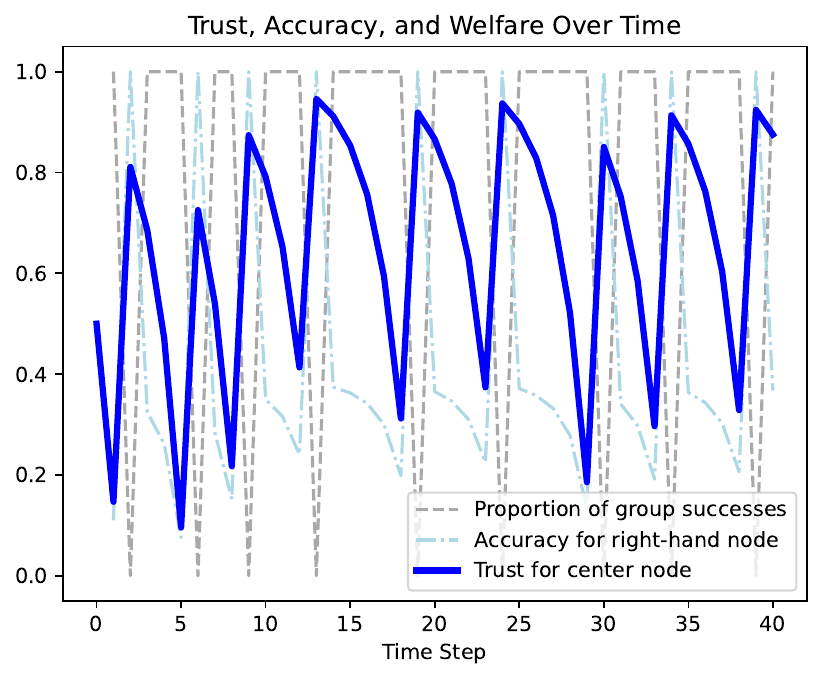}
    \caption{Trust oscillation induced by the optimal welfare maximizer. $T=\frac{2}{3}$ and $\mathcal G$ is a 3-node chain (as seen in Figure~\ref{fig:3-node chain}). 
    Parameters are $c=0.3, r=0.5, B=1, \alpha=0.8, \tau_0=0.5$. The above strategy is optimal for at least $\gamma\in[0.4, 0.5]$ (see Appendix~\ref{app:trust} for details and other $\gamma$).}
    \label{fig:trust-oscillation_medrate}
\end{figure}

An optimal accuracy-maximizer will trivially predict \textit{all-defect}, inducing a stable low-welfare state with perfect accuracy, and trust approaching 100\%. However an optimal welfare-maximizer must handle the tension between accuracy and welfare, resulting from the trust variable. Unless there is a discount factor $\gamma$ that is so high that the predictor indefinitely increases trust for a higher future reward, this will lead to oscillatory behaviour in trust. While a prediction of \textit{all-cooperate} matches the best-response of both left and right nodes, it does not for the center node. To induce \textit{all-cooperate}, the prediction must distribute error in a way that minimally harms accuracy, while maintaining the welfare-maximizing outcome. We derive and prove in Appendix~\ref{app:trust} the optimal welfare-maximizing policy, for different values of $\gamma$. This results in gradually lower confidence for the center node until it is necessary to boost it. For certain $\gamma$'s, the optimal boost is to induce the most surprising outcome for a given $\alpha$, causing an abrupt trust increase (Figure~\ref{fig:trust-oscillation_medrate}).

\section{LEARNED PREDICTOR AND SIMULATIONS}

As the population size $|V|$ grows, analysis becomes more complex. We resort to simulations and learned predictors to study larger systems.

We choose to represent the predictor through a neural network, which receives as input the population's actions in the previous time-step: $\boldsymbol{\hat y_t}=f_{\phi}(a_{t-1}):\{\text{0,1,}\varnothing\}^{|V|}\rightarrow [0,1]^{|V|}$. The loss is either cross-entropy for accuracy maximization, a differentiable proxy of number of successful groups for welfare maximization, or a combination of both. To combine both in a multi-task objective, we  follow the approach of \citet{sener2018multitask} (see Appendix~\ref{app:arch_accuracies}). %
Here we consider a finite horizon of 20 rounds per game with no discounting. A predictor with access to multiple games performs gradient descent after each, assuming access to the inner behaviour of agents. To maximize the number of successful groups, it backpropagates through a differentiable version of their decision rule and of the payoff (Appendix~\ref{app:grad_decomp}).

When optimizing for social welfare, the predictor still needs to consider prediction accuracy in order to maintain agents' trust. Let $\tilde y_{i,t}=\sigma(\pi_{C_i,t}-\pi_{D_i,t})$ be a differentiable proxy of agents' true decision rule $y_{i,t}=\mathds 1[\pi_{C_i,t}-\pi_{D_i,t}>0]$. We analyze here a proxy goal $\hat U_C=\sum^T_{t=1} \sum^N_{i=1} \tilde y_{i,t}$ whose gradient can be decomposed in two components:

\begin{figure}[tbp]
    \centering
    \includegraphics[width=\columnwidth]
    {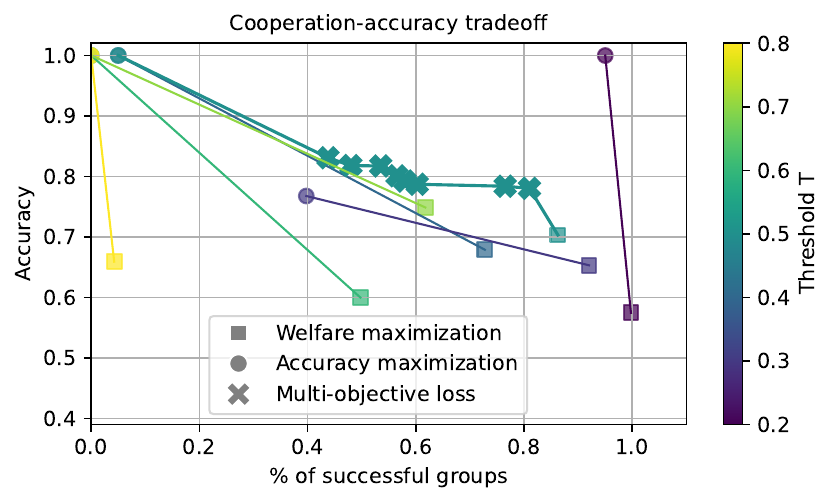}
    \caption{Accuracy vs. social welfare trade-off for different threshold values. Pareto front computed through multi-objective optimization for $T=0.5$. All experiments were conducted using a scale-free $\mathcal{G}$ with 20 nodes and mean degree of 2 \citep{barabasi1999emergence}, $c=0.2, B=1, r=0.4$, $\alpha_{i,j}=0.8$ and $\tau_0=0.5$.}
    \label{fig:tradeoffs}
\end{figure}

\vspace{-.5cm}
\begin{multline}
    \nabla_\phi \hat U_{\text{C}} = \sum^T_{t=1} \sum^N_{i=1}
    \psi_{t,i}(\phi)[ 
    \\(g(T_i|\hat y_{j\in \mathcal{N}(i)}(\phi))
- g(T_i|\alpha_{i,j\in \mathcal{N}(i)})) 
\underbrace{\nabla_\phi\tau_{t,i}(\phi)}_{\text{accuracy}}\\
+\tau_{t,i}(\phi) 
\underbrace{\nabla_\phi g(T_i|\hat y_{j\in \mathcal{N}(i)}(\phi))}_{\text{steering}}]
\end{multline}

$\psi_{t,i}(\phi)=\tilde y_{t,i}(1-\tilde y_{t,i})rB$ is a scalar which is higher when agents are closer to flipping their choice of action between cooperation and defection. An optimizer using this goal needs to control accuracy to keep trust high, and steer towards higher cooperation when trust is high. In practice we use a slightly more complex goal $\hat U_{Pop}$ that is closer to true social welfare, leading to qualitatively similar empirical results and amenable to a similar analysis (Appendix~\ref{app:grad_decomp}).

In Figure~\ref{fig:tradeoffs} we observe the result of training either for accuracy or welfare maximization, for different values of threshold (experimental details in Appendix~\ref{app:arch_accuracies}). 
This extends the theoretical analysis of \S\ref{sec:theoretical_dynamics} to larger populations and more complex $\mathcal G$'s, illustrating that similar patterns hold.
Unless the threshold is very low ($T\in\{0.2, 0.3\}$), a predictor maximizing accuracy will induce states of very low cooperation (find in Appendix~\ref{app:counter_example} a discussion related to the role of $T$ and $\mathcal G$). A predictor maximizing welfare can prevent this, but at the expense of accuracy. This is in line with \S~\ref{sec:simple_envs} where both metrics may be impossible to maximize simultaneously, and with \S~\ref{sec:converge_low_welf} where most initializations lead to low welfare.
To overcome this, we follow \citet{sener2018multitask} to jointly optimize for both metrics. For $T=0.5$, we compute the Pareto front in this way.

\begin{figure*}[t]
    \centering
    \begin{subfigure}{0.32\textwidth}
        \centering
        \includegraphics[width=\linewidth]{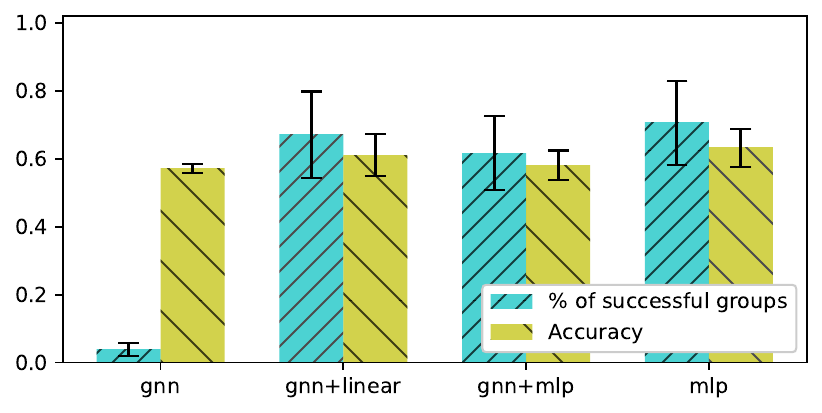}
        \subcaption{$T=0.4$}
        \label{fig:welfare_architectures_t4}
    \end{subfigure}
    \hfill
    \begin{subfigure}{0.32\textwidth}
        \centering
        \includegraphics[width=\linewidth]{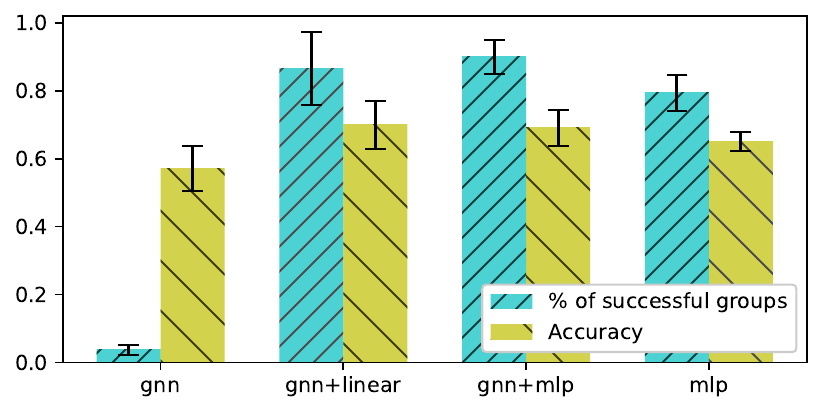}
        \subcaption{$T=0.5$}
        \label{fig:welfare_architectures_t5}
    \end{subfigure}
    \hfill
    \begin{subfigure}{0.32\textwidth}
        \centering
        \includegraphics[width=\linewidth]{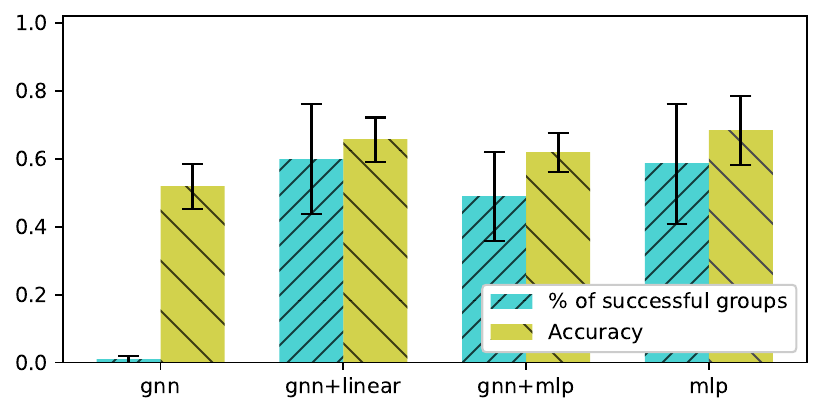}
        \subcaption{$T=0.6$}
        \label{fig:welfare_architectures_t6}
    \end{subfigure}

    \vspace{5pt}

    \caption{Performance of different architectures when optimized to maximize social welfare, for different $T$.}
    \label{fig:welfare_architectures}
\end{figure*}

Regarding architecture choices, we use a multilayer perceptron (MLP), a graph neural network (GNN), GNN+MLP and GNN+linear (Figure~\ref{fig:welfare_architectures}). For an MLP, a concatenation of all nodes' actions is provided as input, and their actions for the next step are jointly predicted. Having a GNN followed by an MLP or a linear layer can provide a gain over MLP alone, by adding information about $\mathcal G$. This pattern, however, does not hold for all $T$'s. Interestingly, GNNs alone,  being the only model unable to do centralized coordination, are not able to promote cooperation. For two equal nodes, some settings may require one to cooperate and the other to defect. A GNN %
is unable to provide different outputs to equal nodes. When optimizing for accuracy, this limitation of GNNs goes by unnoticed (Appendix~\ref{app:arch_accuracies}). We also provide in Appendix~\ref{app:arch_accuracies} an ablation study of the performative gradient used to maximize cooperation. We show the importance of different components, to guide future research in estimating performative gradients dependent on trust.

\section{RELATED WORK}

Machine learning began considering performative effects by assuming explicit models of adaptation \citep{adversarial2004dalvi}. This line of work became known as strategic classification \citep{hardt2016strategic}, initially considering samples with static outcomes $y$ and adaptive features $x$. Later works consider indirect changes in $y$ from causal effects of $x$ on $y$ \citep{miller2020strategic_disguise, horowitz2023causal}. In our setting we predict actions $y$ which agents select to maximize their utility. Predictions are based on past observations of actions, modeling $\mathbb P(Y)$ for single-round games and $\mathbb P(Y_t|Y_{t-1})$ or $\mathbb P(Y_t|Y_{1:t-1})$ for multiple rounds.

The framework of performative prediction \citep{perdomo2020performative} provides theory for a class of adaptation behaviours, showing convergence of RRM, an algorithm which retrains a model after each distribution shift. However they require strong smoothness assumptions not verified in our setting, and do not model explicit interdependencies among predicted. We show, in our setting, how RRM can impact welfare, identifying issues ignored when monitoring accuracy alone.

In existing literature, the utility of predicted agent $i$ depends directly and exclusively on predicted label $\hat y_i$. In our setting utility depends on others' actions, hence only indirectly on predictions. This is a realistic setting which has not been studied before, raising optimization challenges since RRM is not applicable \citep{miller2021outside_echo, izzo2021performative_gradient}. Interestingly, \citet{izzo2021performative_gradient} decompose the performative gradient $\nabla_{\theta} \mathcal L(\theta)$ into an ``easy" component $\nabla_{1} \mathcal L(\theta)$ assuming the distribution is static, and a ``hard" component 
$\nabla_{2} \mathcal L(\theta)= \nabla_{\theta} \mathcal L(\theta) - \nabla_{1} \mathcal L(\theta)$ 
which is the remaining part that requires knowledge of the adaptation. RRM only uses $\nabla_{1} \mathcal L(\theta)$, but maximizing welfare in our setting requires $\nabla_{2} \mathcal L(\theta)$, since $\nabla_{1} \mathcal L(\theta)=0$.

Multi-agent extensions of performative prediction have focused mostly on multiple predictors \citep{li2022multiagent,piliouras2022multi,wang2023network_ppgames}. In \citet{eilat2022strategic_graphnns} predicted outcomes depend on a graph $\mathcal G$ because the classifier assumes it. \citet{mendleranticipating} mention spill-over effects as a way to give a causal treatment to social influence. \citet{hardt2023algorithmic_collective} consider predicted agents that coordinate to influence the training of a classifier. \citet{hardt2022performativepower} propose a lower bound on how much a predictor can influence reality, both for single and multiple predictors. %
\cite{brown2022performative_stateful} consider a stateful version of performative prediction where the previous distribution is enough to define a state, unlike in our setting, where $\boldsymbol{\hat y}_t$ does not contain all information about $\tau_t$. Surprisingly we have not found any existing work on the role of trust in performativity, highlighting a need for future work in this direction.

Regarding alternative goals to accuracy, recommender systems may wish to preserve content diversity \citep{eilat2023performative_recommendation}, and schools can use predictions in an attempt to improve graduation rates \citep{perdomo2023difficult}. 
\citet{vo_krikamol2024causal} consider a trade-off between selecting good candidates and maximizing their improvement, with consequences for agent welfare. \citet{levanon2021practical} balance accuracy and user utility through regularization. \citet{kim2023making} suggests learning a single predictor for many possibly-competing goals.

Behavioural economics has studied cooperation in non-linear social dilemmas \citep{milinski2008collective_risk}, which were modelled using evolutionary game theory \citep{santos2011CRD_collectiveriskdilemma} and used to study mechanism design \citep{gois2019reward}. Another evolutionary model describes adherence to pandemic mitigation measures, showing oscillatory behaviour \citep{glaubitz2020oscillatory}. When predictions are used for downstream tasks, the cost of obtaining a prediction can also be considered in the final goal 
\citep{perdomo2024predictions_value,shirali_hardt2024allocation}.

\section{DISCUSSION}

The proposed setup points to difficult choices that arise in multi-agent performative scenarios. Optimizing for something other than accuracy raises ethical questions, even when the predictor acts in the predicted's interest. However, this work shows that focusing only accuracy is not neutral and can have negative consequences for the predicted. This is an active topic of research in philosophy \citep{van2021pandemic_perform}. \citet{khosrowi2023managing} proposes two opposing views: \textit{mitigation} and \textit{appraisal}. The appraisal view suggests that model evaluation should consider performative effects — e.g. a reduced death toll after a pandemic prediction, and possibly higher accuracy had there been no performativity. However, evaluating models by their consequences undesirably injects moral values into model choice (e.g., prioritizing individual freedom vs. public health). The mitigation view seeks stable, accurate predictions by endogenizing causes of performativity, though this choice may also reflect moral values and has negative impact in our setting. Appraisal corresponds to welfare-maximization and mitigation to accuracy in our model. While this is an open question in philosophy, we emphasize that current machine learning approaches overlook such effects, leading to potential harm — an example would be recommendation systems increasing anxiety to boost engagement. If a predictor does not act in the predicted's interest, it could exploit performative effects for manipulation, which is outside our model's scope. Even then, we argue that making this research open is in the public’s interest, to raise awareness and develop defenses. We hope our results inspire future work on better guidelines and algorithms.

\section{CONCLUSION}
We have introduced a framework to study performative effects under game-theoretic settings on a network of predicted agents. 
We show theoretically how social welfare and accuracy can be in conflict, and empirically compute their Pareto front.
Although accuracy may seem like a way to avoid manipulating reality,
multiple accurate outcomes with different social welfare can be induced when performativity is strong enough.
Ignoring side-effects may be more harmful than considering them, making it inevitable to regard performative prediction (partly) as mechanism design in our examples. This brings connections to active research directions in philosophy, with practical impact in existing systems.
Finally, our model opens up opportunities for future work on the roles of trust, information design, welfare and graphs in performativity. %

\subsubsection*{Acknowledgements}
This research was partially supported by the Canada CIFAR AI Chair Program, by a grant from Samsung Electronics Co., Ldt., by an unrestricted gift from Google, and by a discovery grant from the Natural Sciences and Engineering Research Council of Canada (NSERC). F. P. Santos acknowledges funding by the European Union (ERC, RE-LINK, 101116987). Simon Lacoste-Julien is a CIFAR Associate Fellow in the Learning in Machines \& Brains program. We would like to thank Jose Gallego-Posada for the insightful comments and discussion during the development of this work, leading to the analyses in Section \S~\ref{sec:simple_envs}; also Nir Rosenfeld for helpful discussion which helped structure our motivation; Moritz Hardt for emphasizing trust and oscillation in our model (Section \S~\ref{sec:trust_dyn}); Celestine Mendler-Dünner for pointing the stateful aspect of our model; Ana-Andreea Stoica for the pointer to price of anarchy.

\bibliography{main}

\section*{Checklist}

 \begin{enumerate}

 \item For all models and algorithms presented, check if you include:
 \begin{enumerate}
   \item A clear description of the mathematical setting, assumptions, algorithm, and/or model. [Yes]
   \item An analysis of the properties and complexity (time, space, sample size) of any algorithm. [Not Applicable] Note our focus is not on proposing a new algorithm, but a new setting. However, we analyse the convergence of RRM in this setting, an existing algorithm.
   \item (Optional) Anonymized source code, with specification of all dependencies, including external libraries. [Yes]
 \end{enumerate}

 \item For any theoretical claim, check if you include:
 \begin{enumerate}
   \item Statements of the full set of assumptions of all theoretical results. [Yes]
   \item Complete proofs of all theoretical results. [Yes]
   \item Clear explanations of any assumptions. [Yes]     
 \end{enumerate}

 \item For all figures and tables that present empirical results, check if you include:
 \begin{enumerate}
   \item The code, data, and instructions needed to reproduce the main experimental results (either in the supplemental material or as a URL). [Yes]
   \item All the training details (e.g., data splits, hyperparameters, how they were chosen). [Yes]
   \item A clear definition of the specific measure or statistics and error bars (e.g., with respect to the random seed after running experiments multiple times). [Yes]
   \item A description of the computing infrastructure used. (e.g., type of GPUs, internal cluster, or cloud provider). [Yes]
 \end{enumerate}

 \item If you are using existing assets (e.g., code, data, models) or curating/releasing new assets, check if you include:
 \begin{enumerate}
   \item Citations of the creator If your work uses existing assets. [Not Applicable]
   \item The license information of the assets, if applicable. [Not Applicable]
   \item New assets either in the supplemental material or as a URL, if applicable. [Not Applicable]
   \item Information about consent from data providers/curators. [Not Applicable]
   \item Discussion of sensible content if applicable, e.g., personally identifiable information or offensive content. [Not Applicable]
 \end{enumerate}

 \item If you used crowdsourcing or conducted research with human subjects, check if you include:
 \begin{enumerate}
   \item The full text of instructions given to participants and screenshots. [Not Applicable]
   \item Descriptions of potential participant risks, with links to Institutional Review Board (IRB) approvals if applicable. [Not Applicable]
   \item The estimated hourly wage paid to participants and the total amount spent on participant compensation. [Not Applicable]
 \end{enumerate}

 \end{enumerate}

\clearpage
\raggedbottom
 \onecolumn
\appendix

\aistatstitle{Performative Prediction on Games and Mechanism Design: \\
Supplementary Materials}

\section{BEST RESPONSE}
\label{app:best_response}

An agent's best response selects the action with highest expected payoff, between cooperation and defection. Let $k'_i=\sum_{j\in \mathcal{N}(i)}y_j$ be the number of cooperators in $i$'s group, excluding $i$ itself. The payoff gain of switching from defection to cooperation is

\begin{equation}
  \pi_{C_i}(k'_i+1) - \pi_{D_i}(k'_i) = 
    \begin{cases}
      (r-c)B & \text{if } k'_i=\lceil TM_i\rceil -1\\
      -cB & \text{otherwise}
    \end{cases}       
\end{equation}

In words, $i$ gains $(r-c)B$ from cooperating when it is the last member required to overcome the threshold in its group. It loses $cB$ for any other group configuration. Its best response is then to cooperate when it is ``at the threshold" ($k'_i=\lceil TM_i\rceil -1$) and defect otherwise, as long as $c<r$.

Its expectation of others' actions depends on two competing explanations $g(T_i|\hat y_{j\in\mathcal{N}(i)})$ and $g(T_i|\alpha_{i,j\in\mathcal{N}(i)})$, and the likelihood $\tau_i$ of trusting the first option. Each explanation provides the likelihood $\mathbb P (k'_i=\lceil TM_i\rceil -1)=g(T_i|\theta_{j\in\mathcal{N}(i)})$, by using a Poisson binomial distribution to aggregate individual likelihoods $\theta_j$ of each neighbour to cooperate. It should then cooperate if

\begin{gather}
    \mathbb E_{\tau_i}[\mathbb E_{g(\hat y_{j\in\mathcal{N}(i)})}[\mathbb E_{g(\alpha_{i,j\in\mathcal{N}(i)})}[\pi_{C_i}(k'_i+1) - \pi_{D_i}(k'_i)]]]>0 \,(=) \notag\\
    \tau_i g(T_i|\hat y_{j\in\mathcal{N}(i)})+(1-\tau_i)g(T_i|\alpha_{i,j\in\mathcal{N}(i)})>\frac{c}{r}
\end{gather}

\vspace{10cm}

\section{FULL SUCCESS IS NOT ALWAYS ACHIEVABLE}
\label{app:counter_example}
\begin{figure}[ht]
    \centerline{\includegraphics[width=\columnwidth]{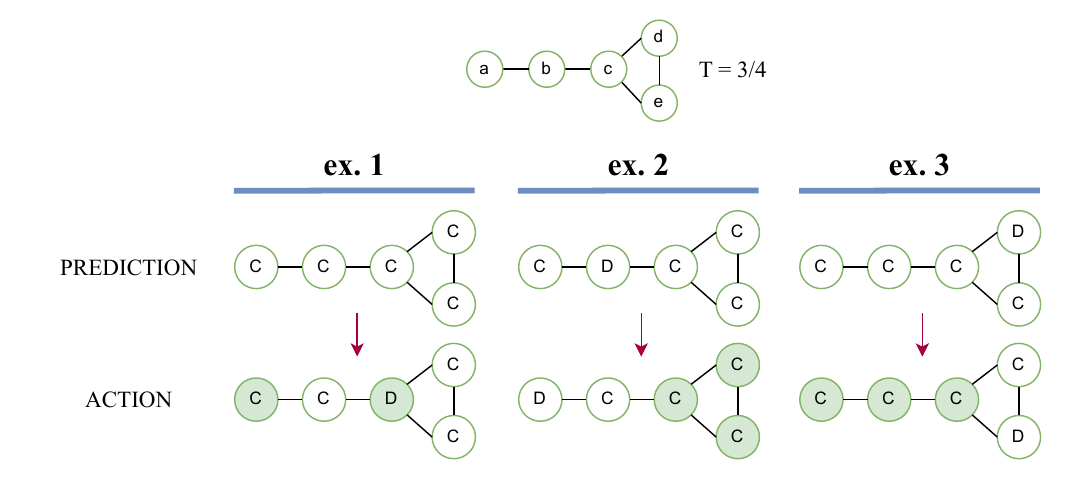}}
    \caption{Achieving full success is not possible for all configurations of $\mathcal{G}$ and $T$. In this counter-example, node $c$ requires $\lceil TM_c\rceil=3$ cooperators out of $M_c=4$, meaning one node in $c$'s group can defect without preventing success. 
    As a consequence $c$ will cooperate only if one of $b,d,e$ is predicted to defect. All the other groups require 100\% of cooperators since they have $M_i<4$ and $T=\frac{3}{4}$. If $c$ doesn't cooperate (ex. 1), it'll prevent success for its neighbours. If any of $b,d,e$ is predicted to defect (ex. 2 and 3), it'll also prevent someone's success. These contradicting requirements make it impossible to reach full success for any given prediction $\boldsymbol{\hat y}$.}
    \label{fig:counter_example}
\end{figure}

Under Assumption~\ref{ass:controllable} there exist combinations of $T$ and $\mathcal{G}$ for which full success is unattainable, even without requiring a self-fulfilling prophecy. This is due to contradicting requirements in neighbour nodes, which cannot be simultaneously satisfied through any prediction $\boldsymbol{\hat y}$. One example is Figure~\ref{fig:counter_example}.

From it, we can formulate sufficient condition for full success to be unattainable, or conversely (through its negation) a necessary condition for full success to be attainable.

\begin{enumerate}
    \item graph $\mathcal{G}$ has a ``hub" node $H$ whose degree $M_H-1$ is higher than any of its neighbours: $\forall i\in\mathcal{N}(H): M_i<M_H$.
    \item Threshold $T\in[0,1]$ is set to $\frac{M_H-1}{M_H}$.
    \item $\forall i\in\mathcal{N}(H), \exists j\in\mathcal{N}(i):M_j<M_H$.
\end{enumerate}

With condition 2, for $H$ to overcome threshold, one out of $M_H$ agents does not need to cooperate (since $M_k\in\mathbb N$ and $\lceil TM_H\rceil=\lceil\frac{M_H-1}{M_H}M_H\rceil=M_H-1$). However, all neighbours $i\in \mathcal{N}(H)$ require 100\% cooperators since they have $M_i<M_H\implies \lceil TM_i\rceil=M_i$. Condition 3 ensures each neighbour of $H$ is connected to another neighbour $j$ with low degree $M_j<M_H$. This combination requires all $i\in\mathcal{N}(H)$ to be predicted to cooperate (i.e. $\forall i\in\mathcal{N}(H), \hat y_i=1$), otherwise their neighbours $j\in\mathcal{N}(i)/\{H\}$ will not cooperate (since they require 100\% cooperators). However, $\forall i\in\mathcal{N}(H), \hat y_i=1\implies y_H=0$ since $H$ can afford one defector in its group. Since $y_H=1$ is a requirement for the success of $i\in\mathcal{N}(H)$, but that requires $\exists! i\in\mathcal{N}(H): \hat y_i=0$, we arrive at contradicting requirements.

This proves Proposition~\ref{prop:nec_success}, since a graph must not obey the condition above, for a prediction to exist which induces full success.

This condition matches empirical observations in Figure~\ref{fig:tradeoffs}. Thresholds that are close to but below 100\% yield low success, even when maximizing welfare. This indicates that there may be no configuration which allows for high success, for settings $(\mathcal{G}, T)$ with high $T$.

\begin{figure}[ht]
    \centerline{\includegraphics[width=.5\columnwidth]{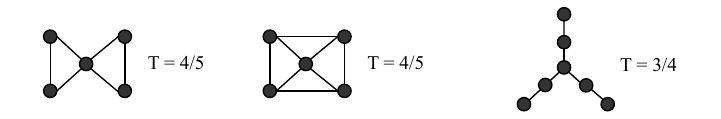}}
    \caption{Other counter-examples where full success is not attainable.}
    \label{fig:more_counter_examples}
\end{figure}

Other counter-examples may be derived from this sufficient condition, such as those in Figure~\ref{fig:more_counter_examples}.

\section{CONVERGENCE TO LOW WELFARE UNDER RRM}
\label{app:low_welfare}

We begin by analyzing the probability of inducing any initial state, from a random initialization of $\boldsymbol{\hat y}_{t=1}$.

\begin{lemma}
\label{lem:initial_prob}
    (Initial state probability) Assume \mbox{$\boldsymbol{\hat y}_1\in\{0,1\}^{|V|}$}, which can be described by $\hat k_1=\sum_i \hat y_{1,i}$ if $\mathcal G$ is fully-connected. Let $n=|V|$, $\hat k_1\sim \text{Bin}(\theta,n)$ and \mbox{$\theta\sim\text{Beta}(\alpha,\beta)$}. The probability of being initialized at a given state $k$ is:
\begin{align}
    \mathbb P(\hat k_1=k) = 
     \frac{n!}{k!(n-k)!}
\frac{\Gamma(k+\alpha)\Gamma(n-k+\beta)}{\Gamma(n+\alpha+\beta)}
\frac{\Gamma(\alpha+\beta)}{\Gamma(\alpha)\Gamma(\beta)}
\end{align}
    
\end{lemma}

\begin{proof}
we have that
\begin{align}
    \mathbb P(\hat k_1=k) &= \int_0^1 \binom{n}{k}\theta^k(1-\theta)^{n-k}\frac{\theta^{\alpha-1}(1-\theta)^{\beta-1}}{\mathrm{B}(\alpha,\beta)}\,d\theta \notag\\
    &=\frac{n!}{k!(n-k)!}\frac{1}{\mathrm{B}(\alpha,\beta)}\int_0^1 \theta^{k+\alpha-1}(1-\theta)^{n-k+\beta-1}\,d\theta
    \notag\\
    &=\frac{n!}{k!(n-k)!}
    \frac{\mathrm{B}(k+\alpha,n-k+\beta)}{\mathrm{B}(\alpha,\beta)}
    \notag\\
    &=\frac{n!}{k!(n-k)!}
    \frac{\Gamma(k+\alpha)\Gamma(n-k+\beta)}{\Gamma(n+\alpha+\beta)}
    \frac{\Gamma(\alpha+\beta)}{\Gamma(\alpha)\Gamma(\beta)} \notag
\end{align}
\end{proof}

\begin{corollary}
    (Probability with increasing population size) under the conditions of \ref{lem:initial_prob}
    \begin{align}
        \displaystyle \lim_{n\to 0}\mathbb P(\hat k_1=k) = 0
    \end{align}
\end{corollary}

\begin{proof}
\begingroup  %
\allowdisplaybreaks
Let $\hat\alpha=\alpha-1$. Using Stirling's formula, we have that $\Gamma(x+\alpha)\sim x^\alpha\Gamma(x)$.
\begin{align}
    \mathbb P(\hat k_1=k) &=
    \frac{n!}{k!(n-k)!}
    \frac{\Gamma(k+1+\hat\alpha)\Gamma(n-k+\beta)}{\Gamma(n+1+\hat\alpha+\beta)}
    \frac{\Gamma(\alpha+\beta)}{\Gamma(\alpha)\Gamma(\beta)}\notag\\
    & \underset{
    \begin{subarray}{c}
    n \to 0 \notag\\
    \text{k=Tn}\notag\\
    \text{n-k=(1-T)n}
    \end{subarray}}{\sim} 
    \frac{n!}{k!(n-k)!}
    \frac{\Gamma(k+1)(k+1)^{\hat\alpha} \Gamma(n-k)(n-k)^{\beta}}{\Gamma(n+1)(n+1)^{\hat\alpha+\beta}}
    \frac{\Gamma(\alpha+\beta)}{\Gamma(\alpha)\Gamma(\beta)}\notag\\
    &=\frac{\Gamma(n-k)}{(n-k)!}
    \frac{(k+1)^{\hat\alpha} (n-k)^{\beta}}{(n+1)^{\hat\alpha+\beta}}
    \frac{\Gamma(\alpha+\beta)}{\Gamma(\alpha)\Gamma(\beta)}\notag\\
    &=\frac{1}{n-k}\frac{(k+1)^{\hat\alpha} (n-k)^{\beta}}{(n+1)^{\hat\alpha+\beta}}
    \frac{\Gamma(\alpha+\beta)}{\Gamma(\alpha)\Gamma(\beta)}\notag\\
    &=\frac{(k+1)^{\hat\alpha} (n-k)^{\beta-1}}{(n+1)^{\hat\alpha+\beta}}
    \frac{\Gamma(\alpha+\beta)}{\Gamma(\alpha)\Gamma(\beta)}\notag\\
    &=\frac{(Tn+1)^{\hat\alpha} ((1-T)n)^{\beta-1}}{(n+1)^{\hat\alpha+\beta}}
    \frac{\Gamma(\alpha+\beta)}{\Gamma(\alpha)\Gamma(\beta)}\notag\\
    &{=\color{blue}\frac{1}{n+1}\frac{(T +\frac{1-T}{n+1})^{\hat\alpha} ((1-T)(1-\frac{1}{n+1}))^{\beta-1}}{T^{\hat\alpha+\beta}}}
    \frac{\Gamma(\alpha+\beta)}{\Gamma(\alpha)\Gamma(\beta)}  \notag\\
    &=\frac{1}{n}\frac{T^{\hat\alpha} (1-T)^{\beta-1}}{T^{\hat\alpha+\beta}}
    \frac{\Gamma(\alpha+\beta)}{\Gamma(\alpha)\Gamma(\beta)} + {\color{blue} O(1/n^2))}\notag\\
    &\to 0 \notag
\end{align}
\endgroup
\end{proof}
By Stirling's formula, we have 
\begin{equation}
    \frac{\Gamma(x+\alpha) }{\Gamma(x)} \sim \frac{(\frac{x+\alpha}{x})^x(x+\alpha)^\alpha}{e^\alpha} =  \frac{(1+\frac{\alpha}{x})^x(x+\alpha)^\alpha}{e^\alpha}\notag
\end{equation}
Now note that $(1+\frac{\alpha}{x})^x = \exp(x \log (1+ \frac{\alpha}{x})) \sim e^\alpha$, and $(x+\alpha)^\alpha = x^\alpha (1+\frac{\alpha}{x})^\alpha \sim x^\alpha$ thus, 
\begin{equation*}
    \frac{\Gamma(x+\alpha) }{\Gamma(x)} \sim x^\alpha
\end{equation*}

We now prove that, for most initializations of $\boldsymbol{\hat y}$ and values of $(N,T)$, RRM converges to a stable point $\mathcal D(\boldsymbol{\hat y}_{\text{Defect}})$ where everyone is predicted to defect and indeed defects, in a self-fulfilling fashion. This leaves the population stuck in a severely suboptimal equilibrium.

Assume a predictor iterates through RRM, predicting a population playing in a fully-connected graph $\mathcal G_f$, with static $\tau=1$. The system becomes a Markov decision problem (MDP) with the following states:

\begin{figure}[!ht]
    \centering
    \includegraphics[width=0.5\linewidth]{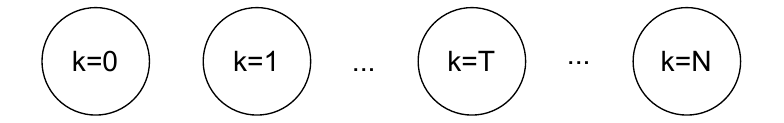}
    \caption{RRM for a single-round CRD with static $\tau=1$ becomes an MDP with N+1 states.}
    \label{fig:enter-label}
\end{figure}

Let $k_i:=\sum_{j\in-i}Y_j$, $k:=\sum_{j\in[1..N]}Y_j$. In the context of this MDP we adopt a slight abuse of notation by using $T$ to denote $\lceil TM_i \rceil$. Since RRM copies the previous set of actions, agents play by best-responding to the previous round. After initializing at some state k, each agent $i$ cooperates if and only if they observe $k_i=T-1$. Note that each agent can only observe $k_i=k$ or $k_i=k-1$, depending on whether $i$ is predicted to defect or cooperate, respectively.

Let us categorize states as:

\begin{enumerate}
    \item $k=T$
    \item $k=T-1$
    \item all others
\end{enumerate}

States 3. move the system into $k=0$, since no agent observes $k_i=T-1$.

For $k=T$, all cooperators $i$ observe $k_i=T-1$, choosing to cooperate next. All defectors $j$ observe $k_j=T$, choosing to defect next. Therefore 1. is a stable point.

For $k=T-1$, only defectors $j$ observe $k_j=T-1$. Therefore defectors and cooperators flip, and 2. leads to $k=N-T+1$. The next step depends on whether $k=N-T+1$ puts us in 1., 2., or 3.:
\begin{enumerate}
    \item If $N-T+1=T(=)N=2T-1$, we arrive in 1. and stabilize at $k=T$.
    \item If $N-T+1=T-1(=)N=2T-2$, we arrive back in 2. meaning we stabilize at $k=T-1$. At each step cooperators and defectors flip, but the count remains constant at $k=\frac{1}{2}$.
    \item Otherwise we are in 3., meaning we arrive at $k=0$ in 2 steps.
\end{enumerate}

At $k=0$:
\begin{itemize}
    \item If $k=0$ is 1. or 3., we stabilize at $k=0$.
    \item If $k=0$ is 2. then $T-1=0(=)T=1$. We cycle between $k=0$ and $k=N$.
\end{itemize}

Therefore we have 1 state converging to $k=T$, or 2 in the particular case of $N=2T-1$. All other states lead to a stable state of $k=0$ in at most 2 steps, with two exceptions: a) with $T=1$, there is a cycle between $k=0$ and $k=N$, and b) with $N=2T-2$ the state $k=T-1$ is also stable. 
This proves theorem~\ref{the:convergence}, where minimizing welfare refers to stabilizing at any state below the threshold, since we consider the cost $cB$ of unnecessarily cooperating is negligible compared to the cost $rB$ of being below $T$.

\begin{figure*}[th]
    \centering
    \includegraphics[width=0.32\textwidth]{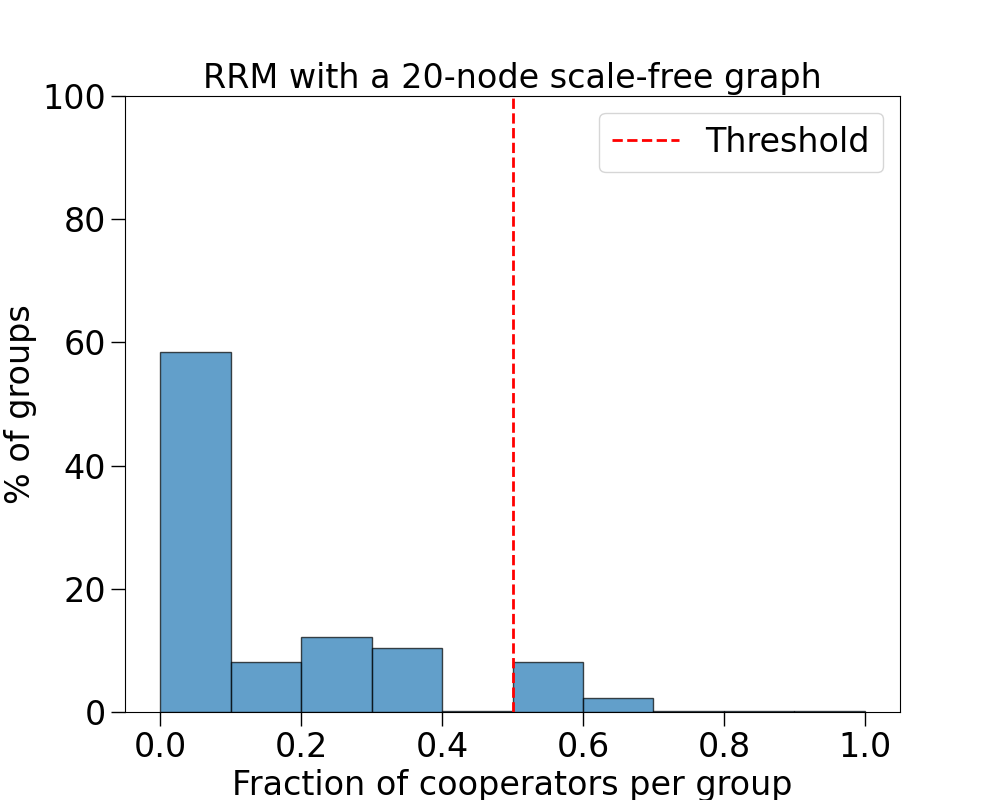}
  \hfill
    \includegraphics[width=0.32\textwidth]{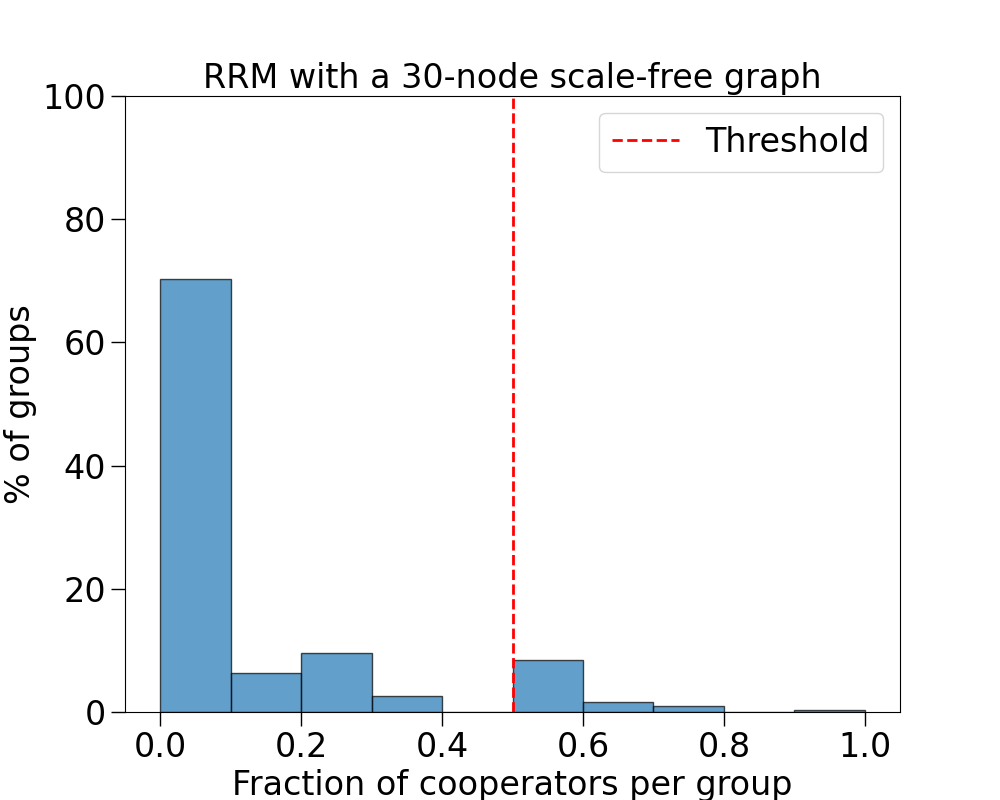}
  \hfill
    \includegraphics[width=0.32\textwidth]{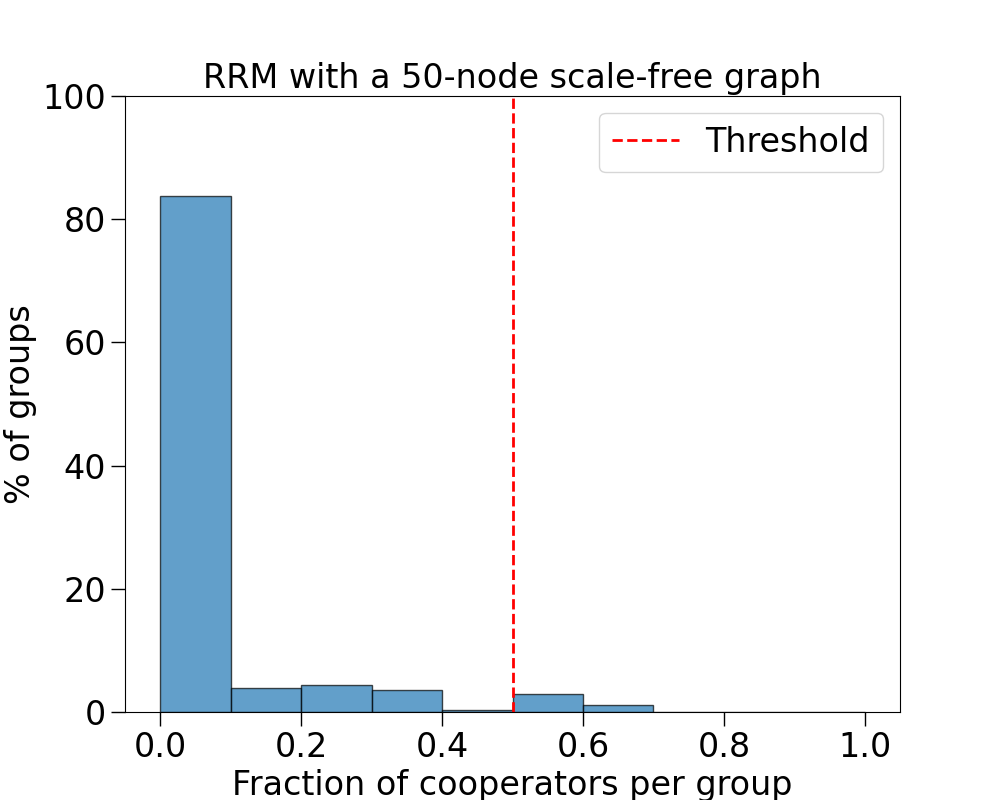}
    \caption{Cooperation per group after applying RRM on a scale-free network with 20 nodes (left), 30 nodes (middle), and 50 nodes (right). Each histogram is the average of 10 different initializations, with a threshold value of 0.5. Each point on the histogram belongs to either a cycle of RRM or a steady state after convergence.}
    \label{fig:temp_rrm_scalefree}
\end{figure*}

Running simulations with scale-free networks, we observe the same pattern which theory predicts for fully-connected graphs — the proportion of time spend below the threshold approaches 100\% as N grows. For runs that cycled instead of converging (which does not happen in a fully connected $\mathcal G$), we report the average proportion of time spent by each group in each state, for one cycle (Figure~\ref{fig:temp_rrm_scalefree}).

\section{CONDITIONS FOR OSCILLATIONS IN TRUST}
\label{app:trust}

\subsection{Setup}

Assume a 3-node chain and threshold $T=\frac{2}{3}$. Let us denote nodes in the chain by 0,1,2 (where 1 is the center node), and actions by $y_{t,i}=1$ (cooperate) and $y_{t,i}=0$ (defect). A predictor maximizing the number of successful groups can, at most, obtain the following rewards per time-step:

\begin{itemize}
    \item Reward $R=3$, if agents play \textit{1—1—1} (all overcome threshold)
    \item Reward $R=2$, if agents play \textit{1—1—0} or \textit{0—1—1} (only nodes 0,1 or 1,2 overcome threshold)
    \item Reward $R=1$, if agents play \textit{1—0—1} (only 1 overcomes threshold)
    \item Reward $R=0$, if agents play for instance \textit{0—0—0} (no one overcomes threshold)
\end{itemize}

$R=3$ is only possible if $y_{t,1}=1$, which requires $\mathbb P(y_{t,0}=0, y_{t,2}=1)+\mathbb P(y_{t,0}=1, y_{t,2}=0)>\frac{c}{r}$. 
We assume this does not happen spontaneously without the external predictor ($\tau_{t,1}=0$). More precisely, we discard values of $\alpha$ and $\frac{c}{r}$ where $2\alpha(1-\alpha)>\frac{c}{r} (=)\frac{1-\sqrt{1-2\frac{c}{r}}}{2}<\alpha<\frac{1+\sqrt{1-2\frac{c}{r}}}{2}$. In this regime, we can only have $R=3$ if $\tau_{t,1}$ is high enough.

Due to symmetry between nodes 0 and 2, obtaining $R=2$ (not show in Figure~\ref{fig:3-node chain}) is not possible with static $\tau=1$, since both observe the same $\hat y_{t,1}$, leading to $y_{t,0}=y_{t,2}$. Even with dynamic $\tau_{t,i}<1$, it is not possible unless $\tau_{0,0}\neq \tau_{0,2}$, which we do not consider here. $R=1$ (with \textit{1—0—1}) requires high enough $\alpha$ or $\tau_{t,0},\tau_{t,2}$, and can be used to boost trust for node 1. 
States with $R=0$ such as \textit{0—0—0} can be used to boost everyone's trust, if needed. 

Note that whether \textit{0—0—0} or \textit{1—0—1} provides the highest increase in $\tau_{t,1}$ depends on how low is the likelihood of the internal model $\mathcal L(\alpha, \boldsymbol{y}_t)=(1-\alpha)^2$ or $\mathcal L(\alpha, \boldsymbol{y}_t)=\alpha^2$, respectively. We assume $\alpha>0.5$, where \textit{0—0—0} provides the highest increase.

\subsection{Optimal Prediction for a Given Outcome}

When inducing a state that requires lowering trust, 
a rational predictor computes the prediction which minimally reduces trust, to avoid future losses. We compute this for $R=1$ and $R=3$ below.

Obtaining \textit{1—0—1} ($R=1$) requires 
a $\hat y_{t,1}$ which induces nodes 0,2 to cooperate, and $\hat y_{t,0}=\hat y_{t,2}=1 \Rightarrow y_{t,1}=0$ (increasing $\tau_{t+1,1}$). Assuming $\tau_{0,0}=\tau_{0,2}$ :

\begin{equation}
\begin{aligned}
\underset{\hat y_{t,1}}{\arg\max} \quad & \mathcal L_0(\hat y_{t,1}, y_{t,1}=0)\\
\textrm{s.t.} \quad & \tau_{t,0}\hat y_{t,1}+ (1-\tau_{t,0})\alpha> \frac{c}{r}\\
  &0 \leq \hat y_{t,1}\leq1    \\
\end{aligned}
\end{equation}

Note $\mathcal L_0(\hat y_{t,1}, y_{t,1}=0)=\mathcal L_2(\hat y_{t,1}, y_{t,1}=0)=1-\hat y_{t,1}$. We can approximate the argmax with $\hat y_{t,1}=\frac{\frac{c}{r}-(1-\tau_{t,0})\alpha}{\tau_{t,0}}+\epsilon$, for a small $\epsilon$. If the solution is unfeasible, $R=1$ is not attainable.

Obtaining $R=3$ requires $\hat y_{t,0},\hat y_{t,2}$ which induce node 1 to cooperate, and $\hat y_{t,1}=1 \Rightarrow y_{t,0}=1,y_{t,2}=1$:

\begin{equation}
\begin{aligned}
\underset{\hat y_{t,0},\hat y_{t,2}}{\arg\max} \quad & \mathcal L_1([\hat y_{t,0},\hat y_{t,2}], [y_{t,0},y_{t,2}]=[1,1])\\
\textrm{s.t.} \quad & \tau_{t,1}(\hat y_{t,0}(1-\hat y_{t,2})+(1-\hat y_{t,0})\hat y_{t,2})+ (1-\tau_{t,1})2*\alpha*(1-\alpha)> \frac{c}{r}\\
  &0 \leq \hat y_{t,0}\leq1    \\
  &0 \leq \hat y_{t,2}\leq1    \\
\end{aligned}
\end{equation}

Note that $\mathcal L_1(\hat y_{t,0},\hat y_{t,2}, y_{t,0}=1,y_{t,2}=0)=\hat y_{t,0}\hat y_{t,2}$. It can be shown that we can approximate the argmax with $\hat y_{t,0}=1-\frac{\frac{c}{r}-(1-\tau_{t,1})2\alpha(1-\alpha)}{\tau_{t,1}}-\epsilon, \hat y_{t,2}=1$, for a small $\epsilon$. If the solution is unfeasible, $R=3$ is not attainable.

\subsection{Existence of Discount Rates for Different Optimal Predictors}

\begin{figure*}[!ht]
    \centering
    \includegraphics[width=.5\columnwidth]{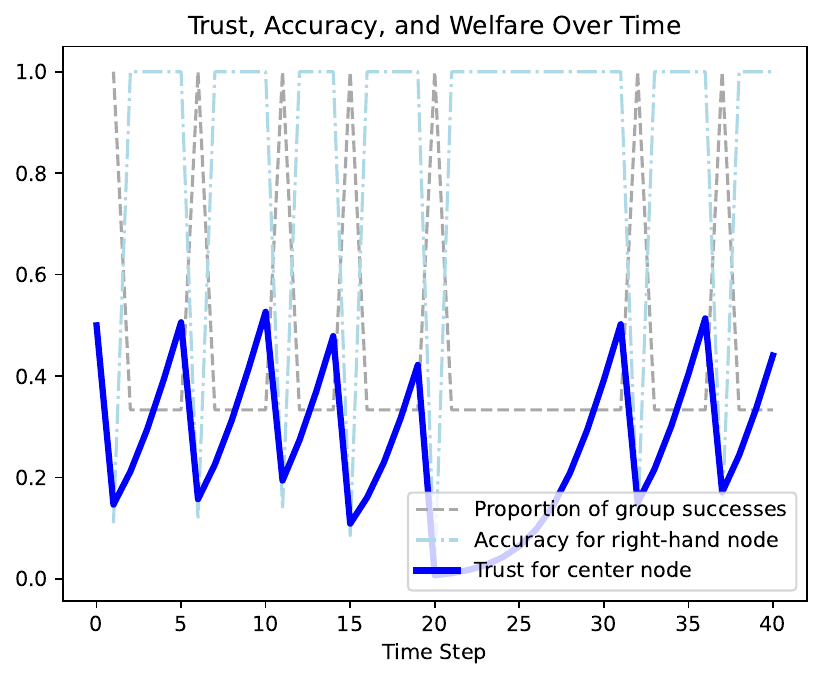}
    \caption{Trust oscillation induced by a welfare maximizer. $T=\frac{2}{3}$ and $\mathcal G$ is a 3-node chain (as seen in Figure~\ref{fig:3-node chain}). Parameters are $c=0.3, r=0.5, B=1, \alpha=0.8, \tau_0=0.5$. The above strategy is optimal for at least $\gamma\in[0, 0.25]$.}
    \label{fig:trust-oscillation_lowrate}
\end{figure*}

Now we assume a predictor which maximizes the number of successful groups, for an infinite horizon with a discount factor of $\gamma$:

\begin{itemize}
    \item Low $\gamma$ (Figure~\ref{fig:trust-oscillation_lowrate}): Consider the scenario where predictor can only obtain $R=1$ (small increase in $\tau_{t,1}$) or $R=0$ (large increase in $\tau_{t,1}$). The predictor prefers to receive $R=1$ now, instead of $R=0$ now and $R=3$ for the next $k$ steps.
    \item Medium $\gamma$ (Figure~\ref{fig:trust-oscillation_medrate}): The predictor prefers to receive $R=3$ now, instead of $R=0$ now and $R=3$ 
    for the next $n$ step. It also prefers to receive $R=0$ now to boost trust and receive $R=3$ for the next $k$ steps, instead of receiving $R=1$ now and $R=1$ for the next $l$-steps, followed by one $R=3$. %
    \item High $\gamma$: The predictor prefers to receive $R=0$ to boost trust and receive $R=3$ for the next $n$ steps, instead of receiving $R=3$ now.
\end{itemize}

Low $\gamma$ occurs for $0\leq\gamma<\kappa$. Consider the extreme case of receiving $R=1$ in $t=0$ followed by $R=0$ forever, alternatively to receiving $R=0$ in $t=0$ followed by $R=3$ forever. We get a lower bound on $\kappa$, dubbed $\kappa_l$, given by $1=\frac{3}{1-\kappa_l}-3(=)\kappa_l=0.25$. For $\gamma<0.25$, predictor picks $R=1$ now even if $R=0$ after (shown in Figure~\ref{fig:trust-oscillation_lowrate}).

Medium $\gamma$ occurs for $\kappa<\gamma<\nu$. Similarly, we get a lower bound $\nu_l$ given by $3=\frac{3}{1-\nu_l}-3(=)\nu_l=0.5$. To show medium $\gamma\in [\kappa,\nu]$ exists, i.e. $\kappa<\nu$, we introduce further assumptions. For the particular values of $r=0.5,c=0.3,\alpha=0.8$ we observe that $l$ is at least 3 (in the best case) and $k$ is at least 2 (in the worst case). Therefore we compare two sequences of rewards: $1, 1, 1, 3, 1, 1, 1, 3, ...$ and $0, 3, 3, 0 , 3, 3,...$. The predictor prefers the latter for approximately $\gamma\geq 0.4$, yielding an upperbound $\kappa_u\approx 0.4$. Therefore, with $r=0.5,c=0.3,\alpha=0.8$, we have a ``medium $\gamma$" at least for $\gamma\in[0.4,0.5]$, whose behaviour is shown in Figure~\ref{fig:trust-oscillation_medrate}.

For higher values of $\gamma$, predictor may prefer $R=0$ followed by $R=3$ $n$-times, instead of $R=3$ now.

We note that, perhaps surprisingly, the posterior computation of trust $\tau_t$ doesn't concentrate around a specific value, as more evidence is gathered. This happens because there is not a static distribution which we try to approximate with $\tau_t$. Instead, whether $\alpha$ or $\boldsymbol{\hat y}_t$ is the correct model varies with time. Additionally, our posterior is over a binary variable, unlike posteriors for continuous variables or categorical with many values, where the tail of the distribution can be reduced over time. This allows for large sudden changes in $\tau_t$, even after several (possibly contradicting) evidence was gathered.

\section{GRADIENT DECOMPOSITION}
\label{app:grad_decomp}

We wish to maximize social welfare $U_{\text{Pop}}=B\sum^T_{t=1} \sum^N_{i=1} ( \mathds 1[\frac{k_i}{M_i}\geq T]*r - y_{i,t}*c)$. Note that $y_i=\mathds 1[\pi_{C_i}-\pi_{D_i}>0]$, meaning there are 2 step-functions $\mathds 1[\cdot]$ in $U_{\text{Pop}}$ where gradient is zero. Both can be replaced by sigmoids $\sigma(\cdot)$, leading to a differentiable approximation $\hat U_{\text{Pop}}$. 
We first analyse a further simplified $\hat U_{\text{C}}$, where the goal is to maximize the total number of cooperators in the population.

\begin{equation}
    U_C=\sum^T_{t=1} \sum^N_{i=1} y_{i,t}=\sum^T_{t=1} \sum^N_{i=1} \mathds 1[\pi_{C_i,t}-\pi_{D_i,t}>0]
\end{equation}

Let $\tilde y_{i,t}=\sigma(\pi_{C_i,t}-\pi_{D_i,t})$ and $\hat U_{\text{C}} = \sum^T_{t=1} \sum^N_{i=1} \tilde y_{i,t}$.

$\nabla_\phi \hat U_{\text{C}} = \sum^T_{t=1} \sum^N_{i=1} \nabla_\phi \tilde y_{i,t}$ %

$= \sum^T_{t=1} \sum^N_{i=1}
\nabla_\phi\sigma(
\underbrace{rB[\tau_{t,i}(\phi) g(T_i|\hat y_{j\in \mathcal{N}(i)}(\phi))+(1-\tau_{t,i}(\phi))g(T_i|\alpha_{i,j\in \mathcal{N}(i)})]-cB}_{h_{t,i}(\phi)}
)$

$=\sum^T_{t=1} \sum^N_{i=1}
\sigma(h_{t,i}(\phi))(1-\sigma(h_{t,i}(\phi)))\nabla_\phi h_{t,i}(\phi)
$

$=\sum^T_{t=1} \sum^N_{i=1}
\underbrace{\tilde y_{i,t}(\phi)(1-\tilde y_{i,t}(\phi)) rB}_{\psi_{t,i}(\phi)}
\nabla_\phi [\tau_{t,i}(\phi) g(T_i|\hat y_{j\in \mathcal{N}(i)}(\phi))
 +  (1-\tau_{t,i}(\phi))g(T_i|\alpha_{i,j\in \mathcal{N}(i)})]
$

$=\sum^T_{t=1} \sum^N_{i=1}
\psi_{t,i}(\phi)
\nabla_\phi [\tau_{t,i}(\phi) (g(T_i|\hat y_{j\in \mathcal{N}(i)}(\phi))
- g(T_i|\alpha_{i,j\in \mathcal{N}(i)}))]
$

$=\sum^T_{t=1} \sum^N_{i=1}
\psi_{t,i}(\phi)
\nabla_\phi [\tau_{t,i}(\phi) (g(T_i|f_{j\in \mathcal{N}(i)}(y^{t-1}_{1:N}(\phi);\phi)) 
- g(T_i|\alpha_{i,j\in \mathcal{N}(i)}))]
$

$=\sum^T_{t=1} \sum^N_{i=1}\psi_{t,i}(\phi)[ (g(T_i|f_{j\in \mathcal{N}(i)}(y^{t-1}_{1:N}(\phi);\phi))
- g(T_i|\alpha_{i,j\in \mathcal{N}(i)})) 
\underbrace{\nabla_\phi\tau_{t,i}(\phi)}_{\text{accuracy}}+
\tau_{t,i}(\phi) 
\underbrace{\nabla_\phi g(T_i|f_{j\in \mathcal{N}(i)}(y^{t-1}_{1:N}(\phi);\phi))}_{\text{steering}}
]
$

$\nabla_\phi\tau_{t,i}(\phi)$ can be interpreted as an accuracy component, where we are interested in having predictions that match past observations in order to increase trust. Interestingly, if the difference $g(T_i|f_{j\in \mathcal{N}(i)}(y^{t-1}_{1:N}(\phi);\phi))
- g(T_i|\alpha_{i,j\in \mathcal{N}(i)})$ becomes negative, it means the model's current predictions are less cooperation-inducing than the agent's innate behaviour. In this case, the gradient will push to decrease accuracy, to incentivize agents to ignore predictions and instead follow their innate behaviour.

The second gradient $\nabla_\phi g(T_i|\hat y_{j\in \mathcal{N}(i)}(\phi))$, or equivalently $\nabla_\phi g(T_i|f_{j\in \mathcal{N}(i)}(y^{t-1}_{1:N}(\phi);\phi))$, can be interpreted as a steering component. If trust $\tau_{t,i}(\phi)$ approaches zero, we won't care about steering since the agents are currently ignoring predictions.

The role of $\psi_{t,i}(\phi)$ is to scale the gradient. Gradients have a larger magnitude when $h_{t,i}(\phi)$ is close to zero, where the agent $i$ is closer to flipping her action between cooperate and defect.

\vspace{.5cm}
Now let success $S_{i,t}=\mathds 1[\frac{k_{i,t}}{M_i}\geq T]$, its differentiable version $\tilde S_{i,t}=\sigma(\frac{k_{i,t}}{M_i}- T)$ and $\hat U_{\text{Pop}}=B\sum^T_{t=1} \sum^N_{i=1} ( \tilde S_{i,t}*r - \tilde y_{i,t}*c)$.

$\nabla_\phi \hat U_{\text{Pop}}=B\sum^T_{t=1} \sum^N_{i=1} ( r*\nabla_\phi \tilde S_{i,t} - c*\nabla_\phi \tilde y_{t,i})$

$\nabla_\phi \tilde S_{i,t} = \tilde S_{i,t}(1- \tilde S_{i,t})\nabla_\phi (\frac{k_{i,t}}{M_i}- T)
= \tilde S_{i,t}(1- \tilde S_{i,t})\frac{1}{M_i} \nabla_\phi k_{i,t}
= \tilde S_{i,t}(1- \tilde S_{i,t})\frac{1}{M_i} \sum_{j\in\mathcal{N}(i)\cup \{i\}} \nabla_\phi \tilde y_{j,t}$

where each $\nabla_\phi \tilde y_{j,t}$ can be analyzed as in $\nabla_\phi\hat U_C$.

Optimizing for either $\hat U_{\text{Pop}}$ or $\hat U_{\text{C}}$ leads to qualitatively similar results empirically.

\pagebreak
\section{EXPERIMENT DETAILS}
\label{app:arch_accuracies}

\subsection{Accuracy Maximization}
\begin{figure}[tbh]
    \centerline{\includegraphics[width=.4\columnwidth]{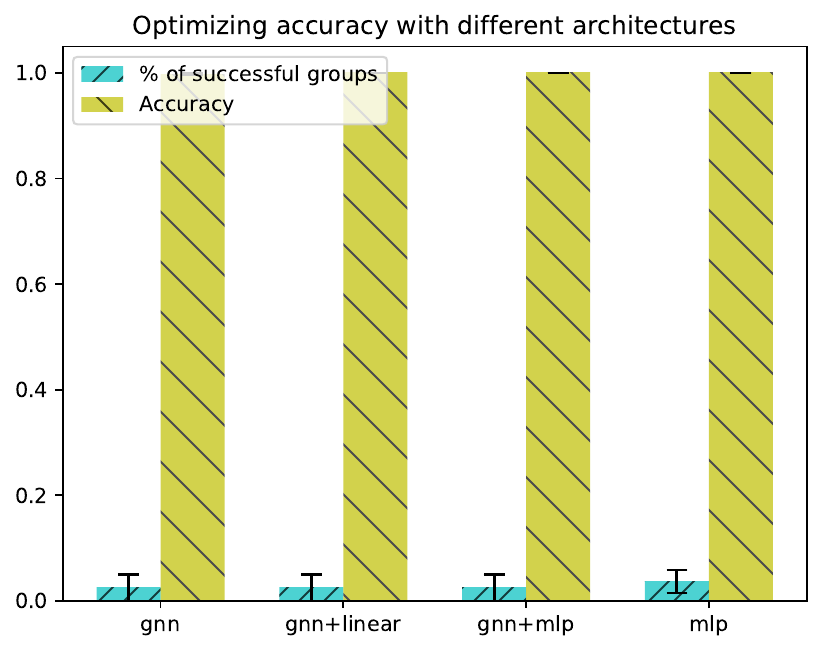}}
    \caption{Performance of different architectures, when optimized to maximize accuracy. All parameters follow Figure~\ref{fig:welfare_architectures}.}
    \label{fig:acc_architectures}
\end{figure}

Any of the used architectures (including a GNN alone) is able to reach perfect accuracy when optimizing for it, at the expense of a very low proportion of successful groups (Figure~\ref{fig:acc_architectures}). These experiments maximize accuracy through RRM. In this setting, by RRM we mean that a) we compute the gradient which makes predictions closer to the observed actions for a game of 20 rounds, b) update the model, c) deploy predictions for a new game, and d) observe new actions. This gradient computation ignores the agents' adaptive behaviour. However, if we assume knowledge of agents' adaptation, we observe empirically that results do not change qualitatively. There is an additional component of the gradient which searches for predictions that push true actions towards the predicted ones. However its effect is negligible.

\subsection{Training, Testing and Pareto Front Computation}

Regarding the computation of the Pareto front, the method proposed in \citet{sener2018multitask} searches for one arbitrary point in the Pareto front. However, due to instability in the training procedure, in our setting we end up traversing a variety of points which do not Pareto dominate each other. The Pareto front displayed in Figure~\ref{fig:tradeoffs} is obtained by training for 200 epochs and keeping only pairs of values (for group success and accuracy) that are not Pareto dominated.

\begin{figure}[b]
    \centering
    \includegraphics[width=0.5\columnwidth]
    {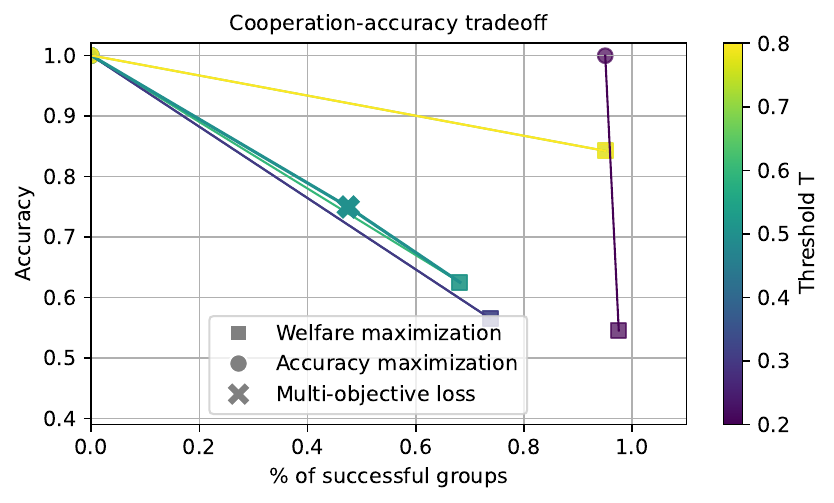}
    \caption{Accuracy vs. social welfare trade-off for different threshold values, setting $\mathcal{G}$ as a lattice with 16 nodes. All other parameters follow Figure~\ref{fig:tradeoffs}.}
    \label{fig:tradeoffs_lattice}
\end{figure}

Since agents' actions are deterministic, we only use one sample during training to compute each gradient step, as well as one sample to evaluate the metric. This sample is a time-series of length 20 (number of rounds per game). We compute the accurary and/or welfare for this sample, and report the max over the 200 epochs. This makes the model trainable on a laptop without the use of a GPU. Future work may consider non-deterministic agent actions by using different assumptions about agent behaviour or about information available to them. For each goal (accuracy, group success, or multi-objective) we compute 3 runs of 200 epochs with different model initializations, and pick the model parameter which yields the highest value. For multi-objective we choose as metric the product of both goals, and display the entire Pareto front containing the point with the highest product.

We provide in Figure~\ref{fig:tradeoffs_lattice} the same experiment but using a different topology for $\mathcal G$, showing a pattern similar to the scale-free $\mathcal{G}$ experiment in Figure~\ref{fig:tradeoffs}.

\subsection{Ablation of Performative Gradient}

\begin{figure}[th]
\begin{center}
\centerline{\includegraphics[width=.8\columnwidth]{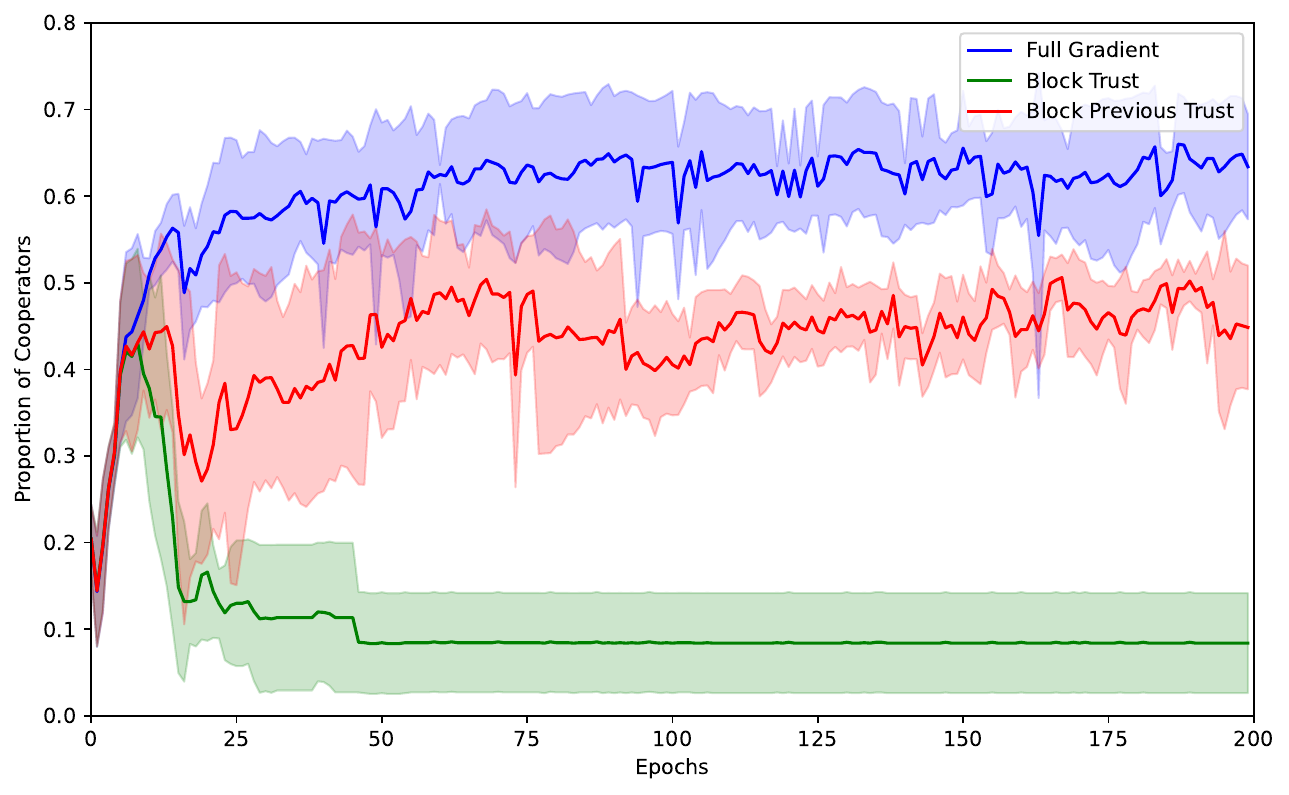}}
\caption{Ablation of cooperation maximizer, excluding components of the performative gradient. Each curve is the average of 5 runs, and the shaded area corresponds to standard deviation.}
\label{fig:ablation}
\end{center}
\end{figure}

Maximizing welfare requires access to the performative gradient, since predictions only influence welfare by changing the distribution. Future work estimating performative gradients under the influence of trust will need to consider which sub-components are most important to estimate. Here we show the importance of different components of the gradient (Figure~\ref{fig:ablation}). 

We consider a surrogate loss for simplicity, which maximizes number of cooperators, instead of number of successful groups. Empirically we observe that maximizing either loss yields similar results. 
We use a differentiable proxy where the predictor assumes each agent's deterministic behaviour ($y_{t,i}= \mathds 1 [ \mathbb E[\pi_{C} - \pi_{D}]>0]$) is replaced by a sigmoid ($\tilde y_{t,i}=\sigma(\mathbb E[\pi_{C} - \pi_{D}]$). %
We then have the following loss per agent $i$, for time-step $t$:

\begin{equation}
    \ell_t(\theta)=-\tilde y_t=-\sigma(\underbrace{\mathbb E[\pi_{C} - \pi_{D}]}_{e_{\tau_t,\theta_t,\alpha}} )
\end{equation}

And the corresponding gradient:

\begin{equation}
    \nabla_\theta\ell_t(\theta)=\nabla_{e_{\tau_t,\theta_t,\alpha}}\ell_t(\theta)\nabla_\theta e_{\tau_t,\theta_t,\alpha}
    =-\sigma(e_{\tau_t,\theta_t,\alpha})\sigma(1-e_{\tau_t,\theta_t,\alpha})\nabla_\theta e_{\tau_t,\theta_t,\alpha}
\end{equation}

Computing $\nabla_{e_{\tau_t,\theta_t,\alpha}}\ell_t(\theta)$ requires knowing $\tau_t$ and $\alpha$, which can be estimated. Computing $\nabla_\theta e_{\tau_t,\theta_t,\alpha}$ however, requires knowing how an agent updates its $\tau_t$, as we show below:

\begin{equation}
    \nabla_\theta e_{\tau_t,\theta_t,\alpha}
    =\nabla_\theta rB[\tau_tg(T_i|\hat y_t)+(1-\tau_t)g(T_i|\alpha)]
    =\nabla_{g(T_i|\hat y_t)}(e_{\tau_t,\theta_t,\alpha})\nabla_\theta(g(T_i|\hat y_t))+\nabla_{\tau_t}(e_{\tau_t,\theta_t,\alpha})\nabla_\theta(\tau_t)
\end{equation}

\begin{itemize}
    \item $\nabla_\theta(g(T_i|\hat y_t))$ does not require knowledge of the agent adaptation.
    \item $\nabla_{g(T_i|\hat y_t)}(e_{\tau_t,\theta_t,\alpha})=rB\tau_t$
    \item $\nabla_{\tau_t}(e_{\tau_t,\theta_t,\alpha})=rB(g(T_i|\hat y_t)-g(T_i|\alpha))$
    \item $\nabla_\theta(\tau_t)=\nabla_{\tau_{t-1}}(\tau_t)\underbrace{\nabla_\theta(\tau_{t-1})}_{\text{recurrence}}+\nabla_{\mathcal L(\hat y_t,y_t)}(\tau_t)\nabla_\theta \mathcal L(\hat y_t,y_t)$
\end{itemize}

To assess the need of knowing the agent adaptation, we set $\nabla_\theta(\tau_t)=0$ (Figure~\ref{fig:ablation}, green curve). Training reaches a peak of around 0.4, as it searches for predictions with increasingly higher likelihood of all agents being at the threshold. However, once we need to balance trust with accuracy, this procedure drops to around 0.1 and does not recover.

In an intermediate scenario, we assume $\nabla_{\tau_{t-1}}(\tau_t)=0$ (Figure~\ref{fig:ablation}, red curve). This assumes knowledge of how current accuracy $\mathcal L(\hat Y_t,Y_t)$ influences current trust $\tau_t$, but not how previous predictions \{$\hat y_{t'}: t'<t$\} influence $\tau_t$ across time. In this setting we can recover after an initial drop in cooperation, unlike with $\nabla_\theta(\tau_t)=0$. However, it does not enable reaching significantly higher values of cooperation. This illustrates the importance of knowing or estimating how predictions affect agents' trust across time.

\clearpage
\raggedbottom

\section{VISUALIZING A POPULATION PLAYING A CRD}

\begin{figure}[ht]
\vskip 0.2in
\begin{center}
\centerline{\includegraphics[width=.5\columnwidth]{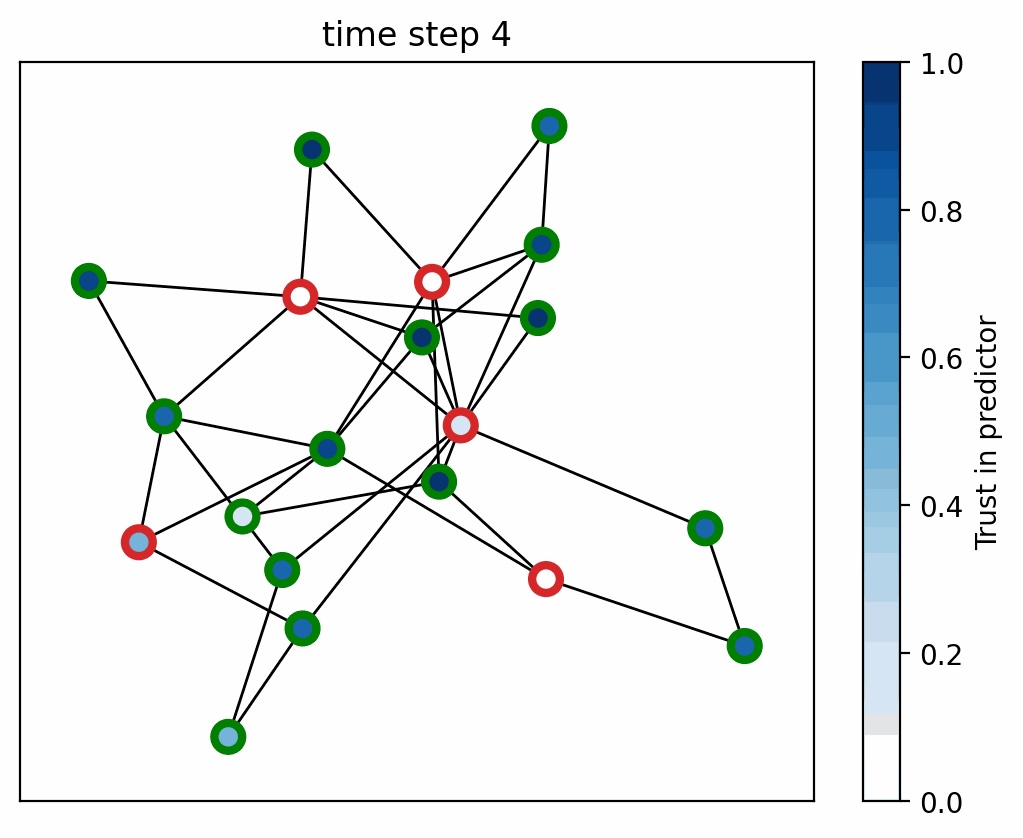}}
\caption{Population playing the Performative Collective Risk Dilemma over a scale-free network \citep{barabasi1999emergence}. Circle borders indicate the agents' last action (green for cooperate, red for defect), and the filling indicates how much the agent currently trusts the predictor.}
\label{fig:population-graph}
\end{center}
\vskip -0.2in
\end{figure}

\end{document}